\documentclass{article} % For LaTeX2e
\usepackage{iclr2025_conference,times}

\usepackage{amsmath,amsfonts,bm}

\def\eqref#1{equation~\ref{#1}}
\def\1{\bm{1}}

\DeclareMathAlphabet{\mathsfit}{\encodingdefault}{\sfdefault}{m}{sl}
\SetMathAlphabet{\mathsfit}{bold}{\encodingdefault}{\sfdefault}{bx}{n}

\DeclareMathOperator*{\argmax}{arg\,max}
\DeclareMathOperator*{\argmin}{arg\,min}

\usepackage{graphicx}
\usepackage{enumitem}
\usepackage{multirow}
\usepackage{colortbl}
\usepackage{booktabs}
\usepackage{amsmath}
\usepackage{amssymb}
\usepackage{xspace}
\usepackage{soul}
\usepackage{comment}
\usepackage{epigraph}
\usepackage{algorithm}
\usepackage{algpseudocode}
\usepackage{tcolorbox}
\usepackage{pifont}

\usepackage{caption}
\usepackage{hyperref}
\usepackage{url}
\usepackage{cleveref}

\definecolor{Gray}{gray}{0.90}
\newcommand{\cmark}{\ding{51}}
\newcommand{\xmark}{\ding{55}}

\def\modelname{MVU\xspace}
\def\llmbaseline{Just-LLM\xspace}
\def\vlmbaseline{Single-Frame-VLM\xspace}

\newcommand{\blue}[1]{\textcolor{blue}{#1}}

\newcommand{\magenta}[1]{\textcolor{magenta}{#1}}

\definecolor{skyblue}{rgb}{0.04,0.40,0.80}
\hypersetup{
    colorlinks,
    linkcolor={red},
    citecolor={skyblue},
    urlcolor={magenta}
}

\definecolor{forestgreen}{rgb}{0.13,0.55,0.13}

\title{Understanding Long Videos with \\ Multimodal Language Models}

\author{Kanchana Ranasinghe, 
Xiang Li,  
Kumara Kahatapitiya \&
Michael S. Ryoo \\
\texttt{kranasinghe@cs.stonybrook.edu}
}

\iclrfinalcopy % Uncomment for camera-ready version, but NOT for submission.
\begin{document}

% There will be a strict upper limit of 10 pages for the main text of the initial submission, with unlimited additional pages for citations.

\maketitle

\begin{abstract}
Large Language Models (LLMs) have allowed recent LLM-based approaches to achieve excellent performance on long-video understanding benchmarks. We investigate how extensive \textit{world knowledge} and strong \textit{reasoning skills} of underlying LLMs influence this strong performance. Surprisingly, we discover that LLM-based approaches can yield surprisingly good accuracy on long-video tasks with limited video information, sometimes even with no \textit{video-specific} information. 
Building on this, we explore injecting video-specific information into an LLM-based framework. We utilize off-the-shelf vision tools to extract three object-centric information modalities from videos, and then leverage natural language as a medium for fusing this information. Our resulting Multimodal Video Understanding (MVU) framework demonstrates state-of-the-art performance across multiple video understanding benchmarks. Strong performance also on robotics domain tasks establishes its strong generality. Code: \href{https://github.com/kahnchana/mvu}{github.com/kahnchana/mvu}
% Our code will be released publicly. 
\end{abstract}

\section{Introduction}
\label{sec:intro}

% discuss these papers in intro @KR
% https://ethz.ch/content/dam/ethz/special-interest/baug/igp/photogrammetry-remote-sensing-dam/documents/pdf/schindler08cvpr.pdf
% https://ai.stanford.edu/~dahuang/papers/cvpr18-fb.pdf

\setlength{\epigraphwidth}{0.9\linewidth}
\epigraph{What can we learn from videos, \\beyond scene context understood from a single natural image?}{}
\vspace{-0.5em}

\noindent
Recent success of large language models (LLMs) and their visual extensions, vision-language models (VLMs), has led to incredible performance on complex language-tied video understanding benchmarks \citep{zhang2023llovi}, particularly on long-video question answering: a task that requires awareness over longer temporal windows \citep{Mangalam2023EgoSchemaAD} as well as causal and temporal action reasoning \citep{dataset_xiao2021nextqa}.
However, the LLMs underlying these approaches contain extensive world knowledge (e.g. understanding of physics, culture, human common sense) and reasoning abilities \citep{yu2023kola,Wang2023GeminiIR}, raising the question of whether they excel at video tasks due to actual \textit{video modality} awareness or simply utilizing world knowledge and contextual information. Such understanding of model reasoning is important for robust deployments avoiding spurious correlation based predictions as well as for better model interpretability \citep{pmlr-v162-yun22a,Xiao2023CanIT}.

In this work, we systematically study this question in the context of video question-answering (QnA) benchmarks, building two modality-constrained baselines to highlight our findings. These two frameworks are tagged \textit{\llmbaseline} and \textit{\vlmbaseline}. 
The first is constrained to access only the task textual query 
% allowing it to answer using only world knowledge 
(i.e. no task-specific visual information). 
The latter is given access to task context with an additional single center-frame from the video as input. 
We discover how these models perform significantly better than random prediction on multiple long-video understanding benchmarks (see \Cref{tbl:baseline}, similar findings in \cite{min2024morevqa}). 
In fact, the latter, utilizing purely world knowledge and contextual information, even outperforms multiple recent state-of-the-art video understanding works (see \Cref{tbl:ego_schema}), challenging the notion of how much \textit{video information} is actually utilized by existing approaches to solve these complex video QnA tasks. 

We next focus on efficient inference to allow rapid experimentation with our LLM based frameworks. Therein, we explore suitable prompting and templating to adapt likelihood selection techniques from prior work \citep{Robinson2022LeveragingLL} to video QnA tasks. 
Our resulting framework achieves more efficient inference with improved performance in comparison to prior work that commonly use auto-regressive generation to tackle long-video QnA benchmarks \citep{zhang2023llovi,Balavzevic2024MemoryCE,wang2025videoagent}. 

Motivated by our initial findings on modality-constrained performance, we study how to inject additional video-specific information into our framework using natural language in a concise and interpretable manner to further improve video understanding. We explore three forms of \textit{object-centric} information modalities, develop pipelines requiring zero video-level training to extract such information using off-the-shelf vision tools, and utilize natural language to fuse this multi-modal information using templating operations. Our resulting approach, termed Multi-Modal Video Understanding (MVU) framework, while achieving state-of-the-art zero-shot performance across long-video understanding benchmarks, also exhibits better interpretability (e.g. exposing video-specific information utilized) through its language-based operation.  Moreover, MVU exhibits generality with its strong performance even on robotics domain tasks (see \textit{zero-shot} robot control in \Cref{app:mvur}). 

\noindent In summary, our key contributions are as follows:
\begin{enumerate}
[leftmargin=3.0em,noitemsep,topsep=-0.4em,itemsep=-1.0ex,partopsep=0ex,parsep=1ex]
    \item Uncover surprisingly strong performance on complex video-language tasks by modality-constrained baselines with limited access to video-specific information. 
    \item Adapting Likelihood Selection strategies to video QnA benchmarks for efficient evaluation.
    \item Novel VLM-based video QnA framework that extracts concise video specific object-centric information followed by natural language based fusion.
\end{enumerate} 
\vspace{0.5em}

We integrate our MVU framework over multiple different baselines and obtain performance improvements across 20 different datasets establishing both its effectiveness and generality. Our evaluations are performed zero-shot with no video-level training on these datasets which cover video QnA tasks (short, medium, and long videos) as well robotics domain tasks. 

% on video QnA benchmarks as well as robotics domain tasks establish its clear benefit. We also highlight how our proposed \modelname is a framework (as opposed to a fixed model) that can easily be integrated with any newer, stronger LLMs or VLMs to utilize their improved capabilities.   

\begin{figure}[t]
    \centering
    \includegraphics[width=\linewidth]{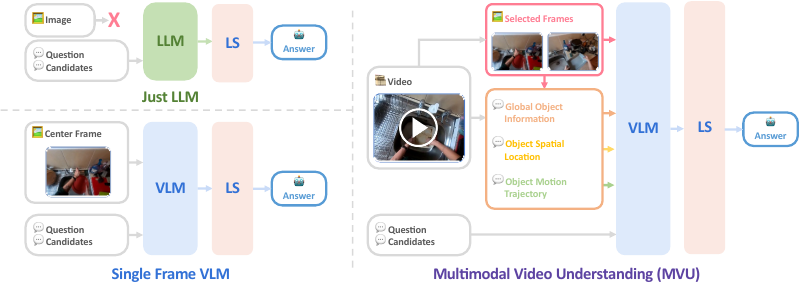}
    \vspace{-1.5em}
    \caption{\textbf{Overview of Framework:}
    We propose three variants of our framework that solves complex long-video question-answering tasks.
    (left-top) \llmbaseline utilizes only world knowledge with zero task-specific awareness. 
    (left-bottom)
    \vlmbaseline processes an additional center frame to obtain task context but accesses no \textit{video} specific information. % (e.g. motion, temporal reasoning). 
    (right) Our complete approach, \modelname extracts three additional object-centric information modalities followed by fusion in language space. LS refers to likelihood selection. 
    % (see \cref{fig:selection}).
    % \kr{refer this fig!}
    % \kk{missing arrows on left-side?} \kk{on right fig, input question in 2 places might be confusing. Can we show this with arrows? ques connects to both lang space and MVU, and there are arrows between inputs and MVU?} 
    % \xl{this figure will be updated, the current one is a draft for concept; but did we mention this figure anywhere?}
    }
    \label{fig:teaser}
    \vspace{-0.5em}
\end{figure}

\section{Related Work}
\label{sec:related}

\noindent \textbf{Video Modality Exploration:}
% Recent rapid progress in visual understanding has motivated 
Multiple recent works dissect the video modality into individual components \citep{pmlr-v162-yun22a,Buch2022RevisitingT,Ranasinghe2021SelfsupervisedVT,Ramasinghe2018CombinedSA}. Single frame baselines are one interesting sub-class \citep{Buch2022RevisitingT,davis1997mei,Zhao_2017_ICCV,8658386,bilen2016dynamic}. Extracting object-centric video modalities is another idea, spanning back to \cite{davis1997mei} which extracts multiple small objects from frames followed by modeling relations across frames and objects. Similarly, \cite{8658386,Zhao_2017_ICCV} combine spatial information with single images to perform video tasks. However, these prior approaches focus on simple video tasks (i.e. action recognition) limited to visual modality. In contrast, our approach tackles the more complex language-tied task of long-video question answering that necessitates strong causal and temporal reasoning over long temporal windows. 
This task is also explored in \cite{Buch2022RevisitingT}, but we differ with likelihood selection, multi-modal information fusion, and usage of modern LLMs.

\noindent \textbf{Long Video Question Answering:}
Long-video question-answering benchmarks are constructed to specifically test strong causal and temporal reasoning \citep{dataset_xiao2021nextqa} over long temporal windows \citep{Mangalam2023EgoSchemaAD}. 
Early works explore querying objects or events based on referential and spatial relations \citep{xu2017msvdqavideo,videoqaaaai2017,Yu2019ActivityNetQAAD}, followed by focus on temporal modeling of sequential events \citep{dataset_lei2018tvqa,lei-etal-2020-tvqaplus,lmappce1_hosseini-etal-2022-knowledge,model_xiao2021hgqa,model_xiao2022vgt}. 
While motivated by these works, \modelname integrates such object information with large language models (LLMs) in a zero-shot manner requiring no video-level training. 
More recent works leverage LLMs \citep{Yu2023SelfChainedIM,papalampidi2023simple,wang2024internvideo2,Balavzevic2024MemoryCE,wang2024tarsier} to directly perform these tasks but require video-caption training. In contrast, our \modelname operates zero-shot on these tasks requiring no video-level training. 
Zero-shot operation is explored in \cite{wang2023vamos,zhang2023llovi,min2024morevqa,Wang2023LifelongMemoryLL,wang2024videotree}, but we differ in using object-centric information modalities and efficient LLM sampling. 

\noindent \textbf{Large Language Model Reasoning:}
Recent LLMs \citep{gpt4,chowdhery2022palm,vicuna2023} demonstrate multiple forms of strong reasoning abilities \citep{Kcman2023CausalRA,Creswell2022FaithfulRU,Liu2023TheMO} including combining different information \citep{Weston2023System2A}. Their recent open-source variants \citep{touvron2023llama2,Anil2023GeminiAF,Jiang2023Mistral7} achieve equally promising skills using scaled-down models \citep{Jiang2023Mistral7} while also demonstrating strong world knowledge \citep{yu2023kola,alkhamissi2024investigating,zhao2023large,Wang2023GeminiIR,xu2024penetrative,li2023language} even in domains such as robotics \citep{li2024llara}. In our work, we leverage these strengths of LLMs for complex video-language tasks, focused on disentangling the effect of their abilities for video QnA tasks. 

\noindent \textbf{Language based Fusion:}
The idea of fusing different modality information using natural language as a medium has been explored in multiple recent works
\citep{Ranasinghe2023LanguagebasedAC,lin2023match,Hanu2022VTCIV,Wang2017AlternativeSR,Hanu2023LanguageAT,zeng2022socratic}. In \cite{Ranasinghe2023LanguagebasedAC,lin2023match}, language is utilized as an implicit medium for self-supervising video action recognition. Multimodal information represented as language is fused with visual information for action recognition and robotics tasks in \citep{Hanu2022VTCIV,Wang2017AlternativeSR,Hanu2023LanguageAT,li2024llara}. We utilize a similar language-as-a-medium fusion of multimodal information, but explore this in the context of complex video-language tasks. 
\cite{zeng2022socratic} is most similar to our work, but we differ with focus on long-video tasks and object-centric information.

\section{Naive Baselines \& Likelihood Selection}
\label{sec:baselines}

In this section, we first establish our problem setting, then discuss adapting likelihood selection for video QnA tasks, and finally introduce two naive LLM based frameworks for video question answering tasks, tagged \textit{\llmbaseline} and \textit{\vlmbaseline} (see \Cref{fig:teaser}). 

\subsection{Problem Formulation}
We focus on two categories of video understanding tasks: 
\begin{enumerate}
	[leftmargin=2.5em,noitemsep,topsep=0.3em,itemsep=-1.0ex,partopsep=0ex,parsep=1ex]
	\item Long Video Question Answering (Multiple-Choice-based Selection)
	\item Open Ended Video Question Answering (Text Generation)
\end{enumerate} 
% \vspace{0.5em}
For the first task, we construct a unified problem formulation accounting their choice based selection aspect. For the latter, we resort to standard LLM based answer generation.  

Consider a video $x_v \in \mathbb{R}^{L\times H\times W\times C}$, a textual question $x_t$, a set of textual candidate answers $Y = \left\{ y_i, i=1, ..., M  \right\}$, and a model $V( \cdot )$ selecting one answer from the given set of answers (noted as $\hat{y}:= V(x_v, x_t, Y)$). Selected $\hat{y}$ should ideally be identical to groundtruth $y_g$.
Here $L, H, W, C$ are the number of frames of the video, frame height, width, and number of channels respectively. $M$ is the number of candidate answers.
For multiple choice based selection tasks, $x_v$, $x_t$, and $Y$ are directly present in dataset.
For N-Way Classification tasks, we set $x_t$ as a generic question (details in \Cref{app:template}) and formulate $Y$ by applying a fixed template to the labels of all N classes of the dataset.
This formulation is used for the remainder of the paper unless a specific exception is noted. 

In the case of open-ended video question answering, we follow standard settings of LLM based text generation for video tasks following \cite{Maaz2023VideoChatGPTTD}.

\subsection{Likelihood Selection}
\label{subsec:likelihood}
The common technique for LLM based  approaches tackling question answering (QnA) tasks is likelihood based choice selection (also referred as Cloze Prompting, see \cite{Robinson2022LeveragingLL}).  Adopting such likelihood based selection for different tasks (or to VLMs) is however not straightforward \citep{Robinson2022LeveragingLL}, leading to most existing long video QnA approaches resorting to LLM based answer generation. In fact, most existing long-video QnA approaches using LLMs / VLMs for choice selection \citep{papalampidi2023simple,wang2022internvideo,Balavzevic2024MemoryCE} resort to full answer generation followed by embedding or template based matching to ground-truth choices, incurring significant inference costs for evaluation. 

\begin{figure}[t]
    \centering
    \includegraphics[width=\linewidth]{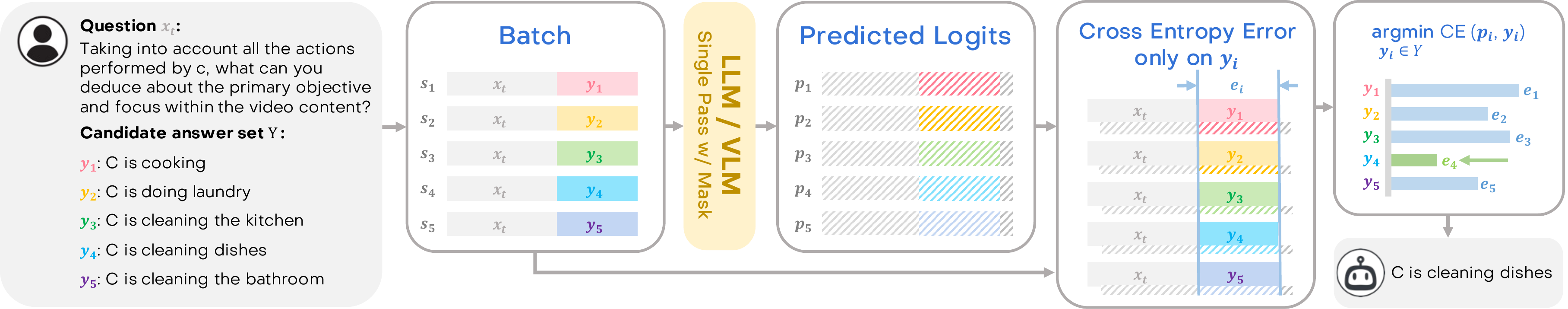}
    \vspace{-1.5em}
    \caption{\textbf{Likelihood Selection Workflow:} 
    We illustrate how the likelihood selection strategy adapted for video QnA tasks can be efficiently parallelized (i.e. calculated with a simple cross-entropy loss in one forward pass, followed by an argmin operation), in contrast to the setting of iteratively generating multiple tokens. 
    }
    \label{fig:selection}
    \vspace{-0.5em}
\end{figure}

In light of this, we explore prompting and likelihood calculation techniques optimal for applying \textit{Likelihood Selection} on long video QnA tasks with either LLMs or even VLMs. 
Adapting this technique unlocks autoregressive LLMs / VLMs ability to solve multiple selection problems with only \textit{one} forward pass as illustrated in \Cref{fig:selection}. This is in contrast to next token sampling requiring iterative generations dependent on previous outputs for each answer token. 
This process uses a likelihood measure based on the LLM latent space allowing better semantic awareness compared to exact or template matching. In addition to the candidate answer batching, we follow prior work to include all candidates in the prompt as well. We direct the reader to \Cref{app:ls_prompts} for complete details on semantic awareness, candidates in prompts, and video QnA specific implementation.

In addition to the considerable inference speed-up from likelihood selection, we also obtain the additional advantages of avoiding LLM hallucinations and deviations from expected output formats over iterative generation strategies applied to similar visual tasks (see \cite{Hanu2023LanguageAT}). We empirically validate the improved performance from such behavior in our ablations (see \Cref{ablate:ls}).

\subsection{Modality Constrained Variants}
\label{subsec:modal_intro}
We next introduce the two modality-constrained variants of our framework tagged \textit{\llmbaseline} and \textit{\vlmbaseline} (illustrated in \Cref{fig:teaser}). 
The former utilizes only the task question injected as language ($x_t$) with no other task-specific information. Note how this naive variant does not access any information extracted from the video for each task instance. 
The latter utilizes an additional center visual frame ($x_v^{c}$), extracted from the center of the video ($x_v$) timeline. This variant accesses no \textit{video-specific} data (e.g. temporal or motion information). The center frame usage ensures no temporal information leakage in frame selection for this variant.

We hypothesize that \llmbaseline with no access to task-specific knowledge is constrained to generate predictions utilizing its internal world knowledge (e.g. physics, culture, human common sense). We refer to this as \textit{world modality}. For a given question regarding a natural video and a set of candidate answers, there is a possibility that one choice is more probable given how our world operates. In cases that this choice turns out to be correct, the internal world knowledge of the LLM allows it to easily select that choice resulting in above random performance. This variant of our framework highlights such cases in long video QnA tasks. 
A similar baseline is used in \cite{min2024morevqa}.

In the case of \vlmbaseline, it is provided with task information but is limited to a single frame, which could possibly provide important scene context. Therein, we refer to this variant as operating with world and \textit{contextual} information modalities. For example, consider a video with a man walking a dog. The scene context of the dog and man combined with the LLM world knowledge and reasoning skills may be sufficient to correctly answer the question with no temporal or motion information. Performance of this variant highlights the prevalence of similar cases in long video QnA tasks when using LLM based approaches.  

We evaluate these two modality-constrained variants and summarize our findings in \Cref{tbl:baseline}. We uncover surprisingly strong performance of both variants on two long-video understanding benchmarks. 
In the case of \llmbaseline variant, we achieve performance significantly higher than random selection (+25.8\% on ES-S / +20.1\% on NextQA-T) using zero visual information. This indicates the large portion of questions in existing video-QnA benchmarks that can be answered correctly purely using world knowledge. 
We also highlight our \vlmbaseline performing on par with state-of-the-art LLM based approach from \cite{zhang2023llovi}. In particular, for ES-S we outperform \cite{zhang2023llovi} which uses information extracted from 180 frames per video incurring an inference cost over 100 times higher than ours. In light of these findings, we argue that long video understanding approaches in particular must focus on learning information beyond what a single frame baseline can achieve, possibly in an interpretable manner. 

\begin{table}[t]
\centering
\small
\begin{minipage}{0.48\textwidth}
\vspace{0.5em}
\caption{\textbf{Modality Constrained Variants:} 
We report accuracy (\%) and inference time per sample (s) on the public subset of EgoSchema (ES-S) and test set of NextQA (NextQA-T) datasets. 
Note that recent state-of-the-art from \cite{zhang2023llovi} (SOTA) and our variants are implemented with common LLMs / VLMs and evaluated under identical settings. 
}
\label{tbl:baseline}
\end{minipage}
\hspace{0.01\textwidth}
\begin{minipage}{0.48\textwidth}
\vspace{-0.5em}
\def\arraystretch{1.0}  % height
\setlength\tabcolsep{0.6em}  % width
\scalebox{0.80}{
\begin{tabular}{lcccccc}
\toprule
\multirow{2}{*}{Method} & \multirow{2}{*}{Param} 
& \multirow{2}{*}{
\begin{tabular}[c]{@{}c@{}}Video\\ Frames\end{tabular}
}
& \multicolumn{2}{c}{ES-S} & \multicolumn{2}{c}{NextQA-T} \\
\cmidrule(lr){4-7}
             &      &   & Acc & Time & Acc  & Time \\ \midrule
Random       & -    & -  & 20.0     &  -      &  20.0    & -          \\ \midrule
\llmbaseline & 7B   & 0  & 45.8     &  0.41   &  40.1    & 0.55       \\ 
SF-VLM       & 13B  & 1  & 55.8     &  1.89   &  51.2    & 2.03       \\ \midrule
SOTA         & 20B  & 180  & 50.8     & 381     &  54.3    & 207        \\ 
\bottomrule
\end{tabular}
}
\end{minipage}
\vspace{-1.0em}
\end{table}

Therein, we introduce \textit{Multimodal Video Understanding} (\modelname), a simple framework that aggregates multimodal video-relevant information in an interpretable manner using natural language and achieves significant improvements over baselines across multiple datasets.

\section{Multimodal Video Understanding Framework}
\label{sec:mvu}
\begin{figure}[t]
    \centering
    \includegraphics[width=\linewidth]{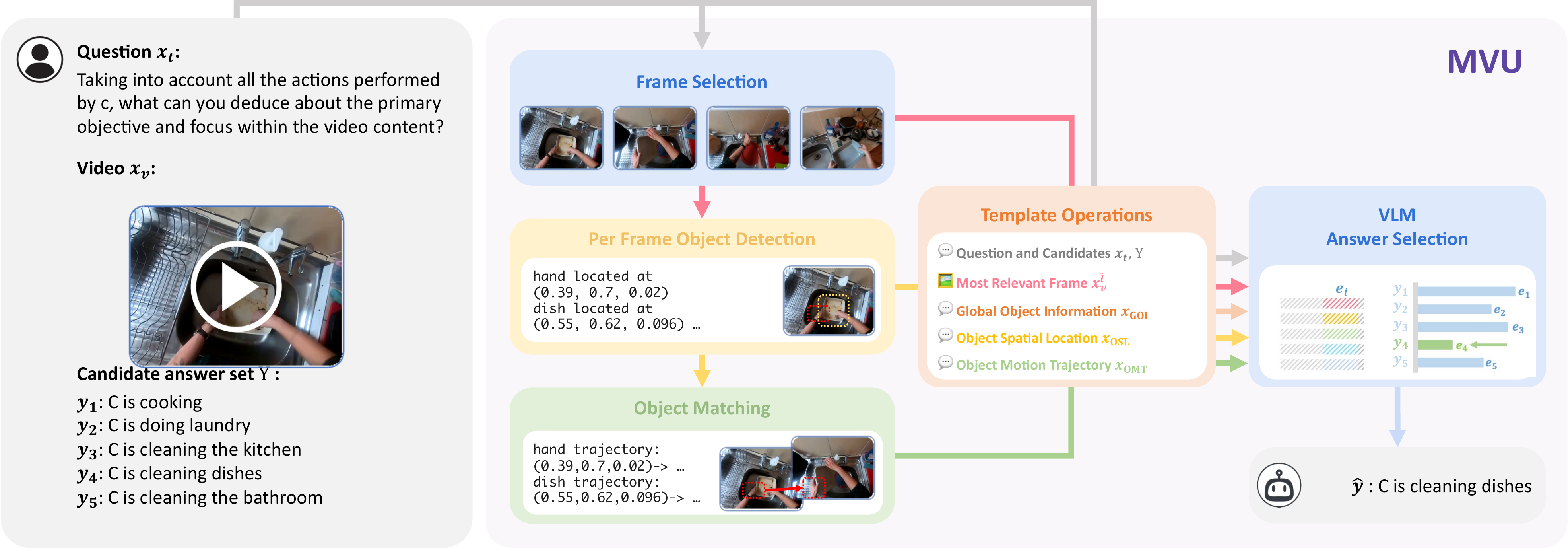}
    \vspace{-1.5em}
    \caption{
    Overview of proposed framework for Multimodal Video Understanding, \modelname. 
    % \textbf{Overview of Architecture:} 
    % Our proposed \modelname framework includes a Frame Selection Module, Object Detector and Tracker, Template Operations Module, and Likelihood Selection Discriminator.   
    % Our model ... TBA: \xl{this is not a paper from Google lol} \kk{this looks nice :)} \xl{thanks, and I would like to make this one as the teaser}
    }
    \label{fig:arch}
    \vspace{-0.5em}
\end{figure}

In this section, we introduce in detail our Multimodal Video Understanding (\modelname) framework that integrates several information modalities extracted from video using \textit{natural language} as a medium for information fusion. Our approach adapts off-the-shelf vision tools to construct a powerful long video understanding agent that requires no additional training on videos.
We first utilize vision tools to extract information relevant to three object-centric modalities from uniformly sampled video frames. Next, we leverage suitable prompt templates to aggregate these as natural language. This video level information is injected into our \vlmbaseline variant providing it with video specific awareness. We illustrate an overview of our framework in \Cref{fig:arch}.

\subsection{Vision Tools for Video Analysis}

Image trained VLMs contain information valuable for video tasks and have been widely used in prior work \citep{zhang2023llovi}. In our proposed framework, we take a step further, exploring more off-the-shelf vision tools trained only on images, in particular object detection and object tracking approaches, in addition to a VLM re-purposed as an image captioner. 

We use an image captioner to identify all unique objects present within a video. For this purpose, we prompt a generative vision language model to list all objects within a given video frame (image) in an open-ended manner. We note how a VLM trained only on images is sufficient for this. In our case, we use a VLM identical to the one in \cite{zhang2023llovi} but applied on significantly less video frames, making our comparisons fair in terms of model size.   

For the case of object detection, we use an open-vocabulary object detector from \cite{Minderer2022SimpleOO} that is trained only on images, and apply it with object category names from captioner to obtain their location information, i.e. image-space coordinates for each unique object. Given the lightweight nature of this detector in comparison to the image captioner, we note how it can be applied more densely (i.e. on more frames) than the captioner without increasing compute demand significantly. Furthermore, the detector acts as a secondary check, grounding the object category names to individual frames, and therein countering any object hallucinations by the captioner. 

Our final tool is an object tracker from \cite{Wang2018FastOO} used to convert our per-frame object detections into motion trajectories spread across the entire video. We feed the tracking algorithm with the locations of each object alongside per-object features extracted from our detector in order to construct motion trajectories for each unique object. 

% We next focus on how these vision tools are utilized to extract object-centric information from a given video. 

\subsection{Object-Centric Information Modalities}
\label{subsec:obj_modal}

Given off-the-shelf tools suitable for extracting information from videos, we next focus on the exact forms of information, i.e. three object-centric information modalities. We consider all object categories across the video, spatial locations of individual object instances, and their movement across time. We define these as follows:

\begin{enumerate}
    \item \textbf{Global Object Information ($x_{\mathrm{GOI}}$):}
    In this stage, we introduce global information that spans beyond a single video frame. For a given video, we first uniformly sample 8 frames. For each of the 8 selected frames, we utilize our image captioner to generate object lists and obtain a set of distinct object categories contained within each frame across the video. 
    \item \textbf{Object Spatial Location ($x_{\mathrm{OSL}}$):}
    Given objects present per video, we utilize our open-vocabulary object detector to localize each object category (from previous stage) on to frame coordinates. Categories not localized by the detector are dropped. Additionally, we utilize similarity of feature vectors for same class objects to track object instances across frames using our tracker. Following prior work \citep{RanLearningtoLoc23}, we calculate average center coordinates and scale value for each object instance across all frames. 
    This results in a set of distinct objects $O$ across the video, $O = \{(o_1, q_1), (o_2, q_2), ... \}$. \
    Here, $o_k$ describes the object category in natural language while $q_k$ contains the x, y coordinates of object center and the scale term (area of minimal object bounding box as a ratio to image size, i.e. box area $\div$ image size). 
    \item \textbf{Object Motion Trajectory ($x_{\mathrm{OMT}}$):}
    Next, we leverage the calculated cross-frame object tracks and compute motion trajectories for each object. This modifies our set of distinct objects, pairing each object $o_k$ with its trajectory ($o_k^1 \rightarrow o_k^2 \rightarrow ... $) across the video frames. We construct an updated set $Z = \{(o_1, q_1^1 \rightarrow q_1^2 \rightarrow ...), (o_2, q_2^1 \rightarrow q_2^2 \rightarrow ...), ... \}$. Intuitively, this information should explicitly capture object motion information. 
\end{enumerate}
We provide further details including examples of each information modality for selected samples (video question pairs) in \Cref{app:template}. 

This pipeline for extracting per-frame information using an image-trained VLM closely resembles prior work such as \citep{zhang2023llovi}. While motivated by such work, we explore the direction of how more fine-grained information could be extracted from videos to solve these tasks more efficiently. 
Given the role of object interactions in defining the various actions and events in videos, we hypothesize that extracting object-centric information (as opposed to generic frame-level descriptions) followed by modeling of their temporal dependencies would provide more concise representations better suited to efficiently solve these tasks.

% We also highlight how the use of vision tools to extract specific forms of information provides an additional level of interpretability to the functions of our overall pipeline. We next discuss how we fuse each of these information modalities using natural language as a medium. 

\subsection{Language based Fusion}
\label{subsec:oc-modalities}
Inspired by \cite{zeng2022socratic}, we construct our overall framework by injecting these three forms of object-centric information into our setup using natural language. 
We represent each modality in a fixed template-based fusion. 
Global object information is represented as a list of category labels, e.g., $x_{\mathrm{GOI}} = $ \{\textit{person, oven, dishwasher, ..., sink}\}. 
Object spatial location modifies this list to include center coordinates ($x, y$) and scale ($s$) where scale is the area percentage occupied by the best-fitting object bounding box. 
For e.g., $x_{\text{OSL}} = $ \{\textit{person located at} (0.2, 0.3, 0.07), ... , \textit{oven located at} (0.8, 0.6, 0.04)\}. 
Finally, object motion trajectories update the list to contain frame-level trajectories, 
e.g., $x_{\text{OMT}} = $ \{\textit{person moving as} [0.2, 0.3, 0.07] $\rightarrow$ [0.2, 0.4, 0.06]  $\rightarrow$ [0.2, 0.6, 0.08], \textit{oven moving as} ... \}. Similar to the examples, information from each object-centric modality is represented in textual form to allow their direct fusion and integration into our framework (as additional language inputs). 
Therein, we describe the resulting setup, our overall framework \modelname as follows, 
\begin{align}
    \hat{y} = \mathcal{F}_{\text{\modelname}}(x_t, {x}_v^c, x_{\text{GOI}}, x_{\text{OSL}}, x_{\text{OMT}})
\end{align}
where ${x}_v^c$ is the center frame extracted from the video $x_v$ (more details in \Cref{app:template}).
In comparison to prior work such as \cite{zhang2023llovi}, we note that our fused information is more concise allowing better utilization of the fixed context length in an LLM (see \Cref{app:llm_context} for more details).

\begin{table}[t]
\centering
\small
\caption{\textbf{Ego-Schema Dataset Evaluation}: 
We report top-1 accuracy (\%) for video question answering on Ego-Schema \citep{Mangalam2023EgoSchemaAD} test set (5031 videos). Our proposed \modelname achieves state-of-the-art performance on this benchmark under \textit{zero-shot operation with no video level training}. We also draw attention to our modality-constrained SF-VLM baseline that achieves surprisingly competitive performance.
% Approaches utilizing proprietary LLMs with significantly higher parameters (e.g. GPT-4) are \textcolor{gray}{de-emphasized}.
}
\label{tbl:ego_schema}
\vspace{-0.5em}
\def\arraystretch{1.0}  % height
\setlength\tabcolsep{1.1em}  % width
\scalebox{0.88}{
\begin{tabular}{lccccc}
\toprule
Method &
  \begin{tabular}[c]{@{}c@{}}Zero\\ Shot\end{tabular} &
  \begin{tabular}[c]{@{}c@{}}Video\\ Training\end{tabular} & 
  \begin{tabular}[c]{@{}c@{}}Closed\\ Model\end{tabular} & 
                            Params &  Full \\ \midrule 
Random Selection                                &   -    &    -   &   -    &  -   & 20.0 \\ \midrule
VIOLET \citep{violet}                           & \cmark & \cmark & \xmark & 198M & 19.9 \\
FrozenBiLM \citep{yang2022frozenblim}           & \cmark & \cmark & \xmark & 1.2B & 26.9 \\
SeViLA \citep{yu2024sevila}                     & \cmark & \cmark & \xmark & 4B   & 22.7 \\ 
mPLUG-Owl \citep{ye2023mplug}                   & \cmark & \cmark & \xmark & 7.2B & 31.1 \\
InternVideo \citep{wang2022internvideo}         & \cmark & \cmark & \xmark & 478M & 32.1 \\
ImageViT \citep{papalampidi2023simple}          & \xmark & \cmark & \xmark & 1B   & 30.9 \\ 
SeViLA+ShortViViT \citep{papalampidi2023simple} & \xmark & \cmark & \xmark & 5B   & 31.3 \\ 
LongViViT \citep{papalampidi2023simple}         & \xmark & \cmark & \xmark & 1B   & 33.3 \\ 
MC-ViT-L \citep{Balavzevic2024MemoryCE}         & \xmark & \cmark & \xmark & 424M & 44.4 \\ 
InternVideo2 \citep{wang2024internvideo2}       & \cmark & \cmark & \xmark & 7B   & 55.8 \\
Tarsier \citep{wang2024tarsier}                 & \cmark & \cmark & \xmark & 7B   & 49.9 \\ 
Tarsier \citep{wang2024tarsier}                 & \cmark & \cmark & \xmark & 34B  & 61.7 \\ 
Vamos \citep{wang2023vamos}                     & \cmark & \xmark & \xmark & 13B  & 36.7 \\ 
LLoVi \citep{zhang2023llovi}                    & \cmark & \xmark & \xmark & 13B  & 33.5 \\
LangRepo \citep{Kahatapitiya2024}               & \cmark & \xmark & \xmark & 12B  & 41.2 \\ 
Vamos \citep{wang2023vamos}                     & \cmark & \xmark & \cmark & 1.8T & 48.3 \\ 
LLoVi \citep{zhang2023llovi}                    & \cmark & \xmark & \cmark & 1.8T & 50.3 \\ 
LifelongMemory \citep{Wang2023LifelongMemoryLL} & \cmark & \xmark & \cmark & 1.8T & 62.4 \\  
MoreVQA \citep{min2024morevqa}                  & \cmark & \xmark & \cmark &  -   & 51.7 \\ 
VideoAgent \citep{wang2025videoagent}           & \cmark & \xmark & \cmark & 1.8T & 54.1 \\  
VideoTree \citep{wang2024videotree}             & \cmark & \xmark & \cmark & 1.8T & 61.1 \\ 
LVNet \citep{Park2024TooMF}                     & \cmark & \xmark & \cmark & 1.8T & 61.1 \\ \midrule 
SF-VLM (ours)                                   & \cmark & \xmark & \xmark & 13B  & 36.4 \\ 
SF-VLM + \modelname (ours)                      & \cmark & \xmark & \xmark & 13B  & 37.6 \\ \rowcolor{Gray}
LVNet + \modelname (ours)                       & \cmark & \xmark & \cmark & 1.8T & 61.3 \\

\bottomrule
\end{tabular}
}
\vspace{-0.5em}
\end{table}

\section{Experiments}
\label{sec:exp}

In this section, we first discuss our experimental setup and datasets. Next, we evaluate \modelname on multiple video question-answering and robotics task benchmarks followed by ablative studies.  

\vspace{0.5em}
\noindent \textbf{Experimental Setup:}
Our proposed \modelname framework and its variants use off-the-shelf models trained on images, thus requiring no re-training of these models. 
For our evaluations, we directly use these models, utilizing two NVIDIA RTX A5000 24GB GPUs for inference.  
We evaluate on two video question answering datasets focused on long-form videos: EgoSchema \citep{Mangalam2023EgoSchemaAD} and NExT-QA \citep{dataset_xiao2021nextqa}.
We also evaluate using a series of robotics datasets from the Open X-Embodiment robotics dataset \citep{open_x_embodiment_rt_x_2023} to test our model generality (more details in \Cref{subsec:robotics}). 
We discuss further details of pretrained models and datasets in \Cref{app:models_data}. 
Also, note that none of the pretrained components of our framework undergo any form of video-level training.

\subsection{Long Video Question Answering}
\label{subsec:eval_longvid}

Long video question answering benchmarks aim to measure causal and temporal reasoning abilities of models over long temporal windows \citep{dataset_xiao2021nextqa,Mangalam2023EgoSchemaAD}. In this section, we evaluate our framework on two benchmark datasets and present our results in \Cref{tbl:ego_schema} and \Cref{tbl:next}. 

On EgoSchema dataset, results reported in \Cref{tbl:ego_schema} demonstrate the state-of-the-art performance of our framework. 
We integrate \modelname over SF-VLM and LVNet \citep{Park2024TooMF} baselines for fair comparison to work operating under different settings. We reiterate how our approach is both zero-shot and requires no video-level training (and our selected baselines are similar). 
In comparison to prior work utilizing open models, our \texttt{SF-VLM+\modelname} achieves clear performance improvements, even out-performing works using video-caption supervision for training \citep{papalampidi2023simple,Balavzevic2024MemoryCE}. Compared to methods utilizing proprietary closed language models extending to trillion parameter scale \citep{zhang2023llovi,wang2023vamos,min2024morevqa,wang2025videoagent}, our \texttt{LVNet+\modelname} variant using similar scale achieves improved performance. 
We also implement several such large-scale approaches under scaled-down common settings as our smaller variant (details in \Cref{app:baselines}), where we again achieve clear performance gains.

\begin{table}[t]
\centering
\small
\caption{\textbf{Next-QA Dataset Evaluation}:
We report top-1 accuracy (\%) for the Next-QA dataset \citep{dataset_xiao2021nextqa}. Our proposed \modelname achieves state-of-the-art results under zero-shot settings with \textit{no video-level training}. 
% We also highlight our competitive performance across the dataset's causal, why, and how sub-categories. 
In table header, ZS stands for zero-shot and VT stands for video level training. 
}
\label{tbl:next}
\vspace{-0.5em}
\def\arraystretch{1.1}  % height
\setlength\tabcolsep{1.1em}  % width
\scalebox{0.88}{
\begin{tabular}{lccccccc}
\toprule
Method & ZS & VT & Params & Cau. & Tem. & Des. & All  \\ \midrule
Random Selection                       & -   & -  &             & 20.0 & 20.0 & 20.0 & 20.0 \\ \midrule
CoVGT \citep{xiao2023covgt}             & \xmark & \cmark & 149M & 58.8 & 57.4 & 69.3 & 60.0 \\
SeViT \citep{kim2023sevit}              & \xmark & \cmark & 215M & - & - & - & 60.6 \\
HiTeA \citep{ye2023hitea}               & \xmark & \cmark & 297M & 62.4 & 58.3 & 75.6 & 63.1 \\ 
InternVideo \citep{wang2022internvideo} & \xmark & \cmark & 478M & 62.5 & 58.5 & 75.8 & 63.2 \\
MC-ViT-L \citep{Balavzevic2024MemoryCE} & \xmark & \cmark & 424M & - & - & - & 65.0 \\
BLIP-2 \citep{li2023blip2}              & \xmark & \cmark & 4B   & 70.1 & 65.2 & 80.1 & 70.1 \\ 
SeViLA \citep{yu2024sevila}             & \xmark & \cmark & 4B   & 74.2 & 69.4 & 81.3 & 73.8 \\ 
LLama-VQA-7B \citep{ko2023llama-vqa}    & \xmark & \cmark & 7B   & 72.7 & 69.2 & 75.8 & 72.0 \\ 
Vamos \citep{wang2023vamos}             & \xmark & \cmark & 7B   & 72.6 & 69.6 & 78.0 & 72.5 \\ \midrule
Just-Ask \citep{yang2021justask}        & \cmark & \cmark & 66M  & 31.8 & 30.4 & 36.0 & 38.4 \\ 
VFC \citep{momeni2023vfc}               & \cmark & \cmark & 164M & 45.4 & 51.6 & 64.1 & 51.5 \\
InternVideo \citep{wang2022internvideo} & \cmark & \cmark & 478M & 43.4 & 48.0 & 65.1 & 49.1 \\
{SeViLA}\citep{yu2024sevila} & {\cmark} & {\cmark}&{4B}&{61.3}&{61.5} & {75.6} & {63.6} \\ 
CaKE-LM \citep{Su2023LanguageMA}& \cmark & \xmark & 2.7B & 35.7 & 35.3 & 36.8 & 34.9 \\
LLoVi \citep{zhang2023llovi}            & \cmark & \xmark & 13B  & 55.6 & 47.9 & 63.2 & 54.3 \\ 
ViperGPT \citep{suris2023vipergpt}       & \cmark & \xmark & 175B & - & -  & - & 60.0 \\
LLoVi \citep{zhang2023llovi} (GPT-4)     & \cmark & \xmark & 1.8T & 69.5 & 61.0 & 75.6 & 67.7 \\
MoreVQA \citep{min2024morevqa}           & \cmark & \xmark & 1.7T & 70.2 & 64.6 & -    & 69.2 \\
VideoAgent \citep{wang2025videoagent}    & \cmark & \xmark & 1.7T & 72.7 & 64.5 & 81.1 & 71.3 \\
VideoTree \citep{wang2024videotree}      & \cmark & \xmark & 1.7T & 75.2 & 67.0 & 81.3 & 73.5 \\
LVNet \citep{Park2024TooMF}              & \cmark & \xmark & 1.8T & 75.0 & 65.5 & 81.5 & 72.9 \\ \midrule
SF-VLM + \modelname (ours)              & \cmark & \xmark & 13B  & 55.7 & 48.2 & 64.2 & 55.4 \\ \rowcolor{Gray}
LVNet + \modelname (ours)               & \cmark & \xmark & 1.8T & 75.2 & 66.8 & 81.3 & 73.3 \\ 
\bottomrule

\end{tabular}}
\vspace{-1.0em}
\end{table}

% MoreVQA \citep{min2024morevqa}           & \cmark & \xmark & 1.7T & 70.2 & 64.6 & -    & 69.2 \\
% VideoAgent \citep{wang2025videoagent}    & \cmark & \xmark & 1.7T & 72.7 & 64.5 & 81.1 & 71.3 \\
% VideoTree \citep{wang2024videotree}      & \cmark & \xmark & 1.7T & 75.2 & 67.0 & 81.3 & 73.5 \\
% LVNet \citep{Park2024TooMF}              & \cmark & \xmark & 1.8T & 75.0 & 65.5 & 81.5 & 72.9 \\ \midrule
% SF-VLM + \modelname (ours)              & \cmark & \xmark & 13B  & 55.7 & 48.2 & 64.2 & 55.4 \\ \rowcolor{Gray}
% LVNet + \modelname (ours)               & \cmark & \xmark & 1.8T & 75.2 & 66.8 & 81.3 & 73.3 \\ 

% forestgreen

% \dht{SeViLA}\cite{yu2024sevila} & \dht{\cmark} & \dht{\cmark}&\dht{4B}&\dht{61.3}&\dht{61.5} & \dht{75.6} & \dht{63.6} \\ 
% ViperGPT \cite{suris2023vipergpt} & 175B & - & -  & - & 60.0 \\
% LLoVi \cite{zhang2023llovi} (GPT-4) & 1.7T & 69.5 & 61.0 & 75.6 & 67.7 \\
% CaKE-LM w HGA \citep{Su2023LanguageMA}  & \cmark & \xmark & 2.7B & 35.3 & 34.7 & 36.9 & 34.8 \\

Next, we evaluate our framework on the NextQA benchmark and report these results in \Cref{tbl:next}. 
We similarly integrate \modelname with two baselines.
Our \modelname achieves state-of-the-art results under zero-shot settings. While \cite{yu2024sevila} outperforms our approach, we note how they require video-caption localization pretraining and appears to overfit to this dataset considering their relatively lower performance on other datasets (see \Cref{tbl:ego_schema}). 

We also evaluate \modelname on the LongVideoBench dataset which contains even longer videos and present these results in \Cref{app:longvideobench}.
While these three datasets focus on MCQ style QnA, we also explore the generality of our \modelname framework on open-ended style QnA tasks in \Cref{app:open_qa}.

\subsection{Robotics Domain}
\label{subsec:robotics}

\begin{figure}[t]
\begin{minipage}{0.69\textwidth}
    \centering
    \begin{minipage}{\linewidth}
        \centering
        \includegraphics[width=1.0\linewidth]{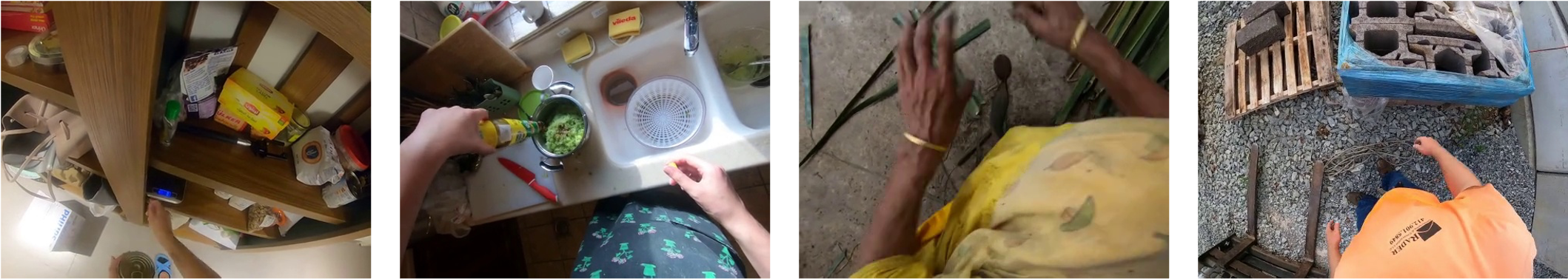}     
    \end{minipage}
    \begin{minipage}{\linewidth}
        \centering
        \includegraphics[width=1.0\linewidth]{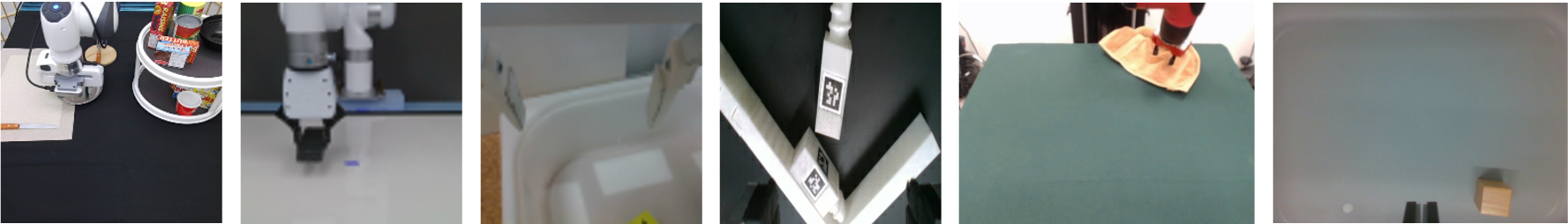}
    \end{minipage}
\end{minipage}
\hspace{0.01\textwidth}
\begin{minipage}{0.29\textwidth}
    \caption{\small
    \textbf{Data Visualization:}
    Example video frames from EgoSchema (top) vs OpenX (bottom) datasets. 
    Robotics domain videos (bottom) appear out of distribution given their controlled environment and robot movements.
    }
    \label{fig:dt_example}
\end{minipage} 
\vspace{-1.0em}
\end{figure}

\noindent \textbf{Action Recognition:}
We investigate generalization capabilities of our proposed \modelname by evaluating across datasets from robotics domain Open X-Embodiment, following a QnA style formulation of the dataset (details in \Cref{app:openx}). We highlight visual differences of this data in \Cref{fig:dt_example}.
We present evaluations in \Cref{tbl:robotics}, which indicate clear performance improvements for \modelname over the baseline from \cite{zhang2023llovi}. The purpose of this experiment is to evaluate the generality of our approach to video domains different from everyday natural videos. We take these promising results to indicate the strong generality of our framework. 
Furthermore, we note how our modality constrained variants do not perform significantly better than random on robotics domain tasks (details in \Cref{app:modality}). We attribute this to significant domain shift in terms of the world of operation.
% (i.e. robotics tasks tend to involve controlled environments very different to what humans face on an everyday basis).   

\noindent \textbf{Robot Control:}
MVU extension for robot control in \Cref{app:mvur} further highlights our generality.

\begin{table}[t]
\centering
\small
\caption{
\textbf{OpenX Detailed Results:}
We report accuracy (\%) for the VideoQA formulation of Open X-Embodiment benchmark. \modelname achieves clear improvements over random selection and LLoVi baseline \citep{zhang2023llovi}. 
In table header, Obs. (observation), size, CC (class count) stand for camera used, number of videos, and number of unique language instructions per dataset, respectively. In observation column, T stands for third-person view (stationary camera that does not move with robot), while F denotes first-person view where camera is mounted on moving robot. Note that total is average weighted by dataset size.  
}
\label{tbl:robotics}
\vspace{-0.5em}
\def\arraystretch{1.1}  % height
\setlength\tabcolsep{1.3em}  % width
\scalebox{0.85}{
\begin{tabular}{lcccccc}
\toprule
Dataset                         & Obs. & Size  & CC  & Random & Baseline & MVU  \\ \midrule
ASU TableTop Manipulation       & T    & 110   & 83  & 13.6   & 19.1   & \textbf{20.9} \\
Berkeley MVP Data               & F    & 480   & 6   & 20     & 26.0   & \textbf{33.1} \\
Berkeley RPT Data               & F    & 908   & 4   & 24.6   & 23.1   & \textbf{26.2} \\
CMU Play Fusion                 & T    & 576   & 44  & 20.3   & 34.0   & \textbf{35.6} \\
CMU Stretch                     & T    & 135   & 5   & 23     & 18.5   & \textbf{24.4} \\
Furniture Bench                 & T    & 5100  & 9   & 20.2   & 24.8   & \textbf{26.4} \\
Furniture Bench                 & F    & 5100  & 9   & 20.2   & 22.6   & \textbf{24.9} \\
CMU Franka Pick-Insert Data     & T    & 631   & 7   & 18.7   & 19.3   & \textbf{21.2} \\
CMU Franka Pick-Insert Data     & F    & 631   & 7   & 23.1   & \textbf{57.8}   & 49.3 \\
Imperial F Cam                  & T    & 170   & 17  & 20     & 22.9   & \textbf{24.1} \\
Imperial F Cam                  & F    & 170   & 17  & 23.5   & 20.6   & \textbf{24.7} \\
USC Jaco Play                   & T    & 1085  & 89  & 21.8   & 26.4   & \textbf{30.6} \\
USC Jaco Play                   & F    & 1085  & 89  & 19.4   & 28.6   & \textbf{32.4} \\
NYU ROT                         & T    & 14    & 12  & 21.4   & \textbf{57.1}   & \textbf{57.1} \\
Roboturk                        & T    & 1959  & 3   & 34.7   & 43.0   & \textbf{44.2} \\
Stanford HYDRA                  & T    & 570   & 3   & 35.1   & 54.7   & \textbf{68.2} \\
Stanford HYDRA                  & F    & 570   & 3   & 31.2   & 45.3   & \textbf{48.9} \\
Freiburg Franka Play            & F    & 3603  & 406 & 20.4   & \textbf{32.2}   & 31.6 \\
Freiburg Franka Play            & T    & 3603  & 406 & 19.7   & 21.8   & \textbf{24.0} \\
LSMO Dataset                    & T    & 50    & 2   & 34.0   & 68.0   & \textbf{72.0} \\
UCSD Kitchen                    & T    & 150   & 8   & 19.3   & 32.0   & \textbf{32.7} \\
Austin VIOLA                    & T    & 150   & 3   & 26.7   & 32.7   & \textbf{33.3} \\
Austin VIOLA                    & F    & 150   & 3   & 30.0   & 33.3   & \textbf{34.0} \\ \midrule
Total                           & -    & 27000 & -   & 22.1   & 28.5   & \textbf{30.4} \\
\bottomrule
\end{tabular}
}
\vspace{-1.0em}
\end{table}

\subsection{Ablations}
\label{subsec:ablate}
In this section, we systematically dissect our overall \modelname framework to establish the usefulness of each of its individual component %
(see \Cref{app:ablate_more} for more ablations). We first ablate our three different information modalities and report these results in \Cref{ablate:modal}. Our evaluations indicate clear performance improvements from each of our object-centric information modalities. 

\begin{table}[t]
\centering
\small
\begin{minipage}{0.48\textwidth}
\caption{\textbf{\modelname Ablation:} 
We report accuracy (\%) on public subset of EgoSchema (ES-S). In table header, VI stand for visual inputs and GOI, OSL, OMT refer to our object-centric information modalities (see \Cref{subsec:obj_modal}). 
Each information modality results in clear performance improvements with our full \modelname achieving best performance.}
\label{ablate:modal}
\end{minipage}
\hspace{0.01\textwidth}
\begin{minipage}{0.48\textwidth}
\vspace{-1.0em}
\def\arraystretch{1.0}  % height
\setlength\tabcolsep{0.6em}  % width
\scalebox{0.88}{
\begin{tabular}{lcccccc}
\toprule
Method         & VI  & GOI & OSL & OMT & ES-S       \\ \midrule
\llmbaseline (ours)    & \xmark & \xmark & \xmark & \xmark & 45.8 \\
SF-VLM (ours)          & \cmark & \xmark & \xmark & \xmark & 55.8 \\
\modelname (ours)      & \cmark & \cmark & \xmark & \xmark & 56.4 \\
\modelname (ours)      & \cmark & \cmark & \cmark & \xmark & 58.6 \\ \rowcolor{Gray}
\modelname (ours-full) & \cmark & \cmark & \cmark & \cmark & \textbf{60.3} \\ \bottomrule
\end{tabular}}
\end{minipage}
\vspace{-1.5em}
\end{table} 

We next perform ablations on our adaptation of likelihood selection strategy for video QnA tasks using Ego-Schema subset (ES-S) and Next-QA test-set (NQA-T) . These results reported in \Cref{ablate:ls} indicate clear performance boosts due to our adaptation of likelihood selection (LS). When removing LS, standard generation (i.e. generate an answer and match against ground-truth selection following \cite{zhang2023llovi}) is utilized with our \modelname framework. We also report naive adaptation of LS following \cite{Robinson2022LeveragingLL} where the choice options are directly used, highlighting the importance of our prompting techniques.  
We also note the accuracy gains obtained through LS, and attribute these to reduced LLM output hallucination and deviations from expected output formats, that are commonly observed with iterative generation \citep{Hanu2023LanguageAT}.

We next ablate our overall framework against the existing work, \cite{zhang2023llovi}, by replacing our \modelname object-centric information pipeline with the frame description approach in \cite{zhang2023llovi}. We construct a setup identical to our framework except for the inputs to our final stage VLM replaced with frame level descriptions. These results reported in \Cref{ablate:baseline} indicate the clear significance and improvement of our proposed object-centric information pipeline over simple frame descriptions. The 8 frame variant is the same speed comparison as \modelname uses captioner only on 8 frames. Our MVU outperforms both that and the slower 16 frame baseline. We also note the performance drop in the baseline when increasing the number of frames from 16 to 180. While consistent with observations in prior works for long-video tasks \citep{Mangalam2023EgoSchemaAD}, we attribute this drop to decreased signal-to-noise ratio with the introduction of additional frame descriptions. This further highlights the importance of selecting useful information from video frames, and we reiterate how the object-centric information in our \modelname framework serves this purpose. 

\begin{table}[t]
\begin{minipage}{0.48\textwidth}
    \centering
    \small
    \caption{
    \textbf{Likelihood Selection (LS) Ablation:} Results indicate clear improvements in both accuracy (\%) and inference time (s) with our adaptation of likelihood selection for video tasks.     
    }
    \label{ablate:ls}
    \vspace{-0.5em}
    \def\arraystretch{1.1}  % height
    \setlength\tabcolsep{0.9em}  % width
    \scalebox{0.85}{
    \begin{tabular}{lcccc}
    \toprule
    Method            & LS & ES-S & NQA-T & Time \\ \midrule
    Generation        & \xmark & 56.4 & 55.3 & 12.7 \\ 
    LS-Naive          & \xmark & 58.2 & 35.8 & 2.42 \\ \rowcolor{Gray}
    LS-MVU (ours)     & \cmark & \textbf{60.3} & \textbf{55.4} & \textbf{2.42} \\ \bottomrule
    \end{tabular}
    }
\end{minipage}
\hspace{0.01\textwidth}
\begin{minipage}{0.48\textwidth}
    \centering
    \small
    \caption{\textbf{Baseline Ablation:} 
    We replace information input to final stage VLM with frame descriptions following \cite{zhang2023llovi}. Accuracy (\%) on public subset of EgoSchema (ES-S). Time in seconds (s).
    }
    \label{ablate:baseline}
    \vspace{-1.0em}
    \def\arraystretch{1.0}  % height
    \setlength\tabcolsep{1.2em}  % width
    \scalebox{0.85}{
    \begin{tabular}{lcccc}
    \toprule
    Method            & Frames & ES-S & Time \\ \midrule
    Baseline          &   180  & 55.4 & 207 \\ 
    Baseline          &    8   & 55.8 & 2.38 \\  
    Baseline          &   16   & 56.2 & 4.72 \\  \rowcolor{Gray}
    \modelname (ours) &   16   & \textbf{60.3} & 2.42 \\ \bottomrule
    \end{tabular}
    }
\end{minipage}
\vspace{-0.5em}
\end{table}

\section{Conclusion}
\label{sec:conc}

% modality constrained variants - findings
% LS - adopt to video tasks, use with MLMs for the first time
% MVU contribution

In this work, we present a multimodal video understanding framework, termed \modelname, that achieves state-of-the-art performance on complex video understanding tasks. In particular, evaluations on long-video question answering and robotics domain question answering demonstrate the strong performance of our \modelname framework as well as its generality. We also adapt likelihood selection for efficient LLM-based answer choice selection, separate video-specific information into three object-centric modalities, demonstrate automated extraction of such information using off-the-shelf vision tools, and propose language-based fusion of this multimodal information. 

We also presented two modality-constrained baselines that uncover surprising insights relevant to LLM based video QnA which serves as a basis for our subsequent \modelname framework. Furthermore, these results highlight the need for careful evaluation of LLM-based video QnA approaches. 
Revisiting our original motivation on \textit{``what we can learn from videos, beyond scene context understood from a single natural image''}, in this work our two modality-constrained variants uncover surprising insights relevant to this question. We first achieve strong results on long-video understanding benchmarks using no video-specific data, and build over that baseline to showcase the additional performance gains achievable through injecting video-specific information. 

\section*{Reproducibility Statement}
Our method utilizes multiple pretrained models, all of which are open-source with model weights freely available. We use the versions of these models hosted on HuggingFace \url{https://huggingface.co} for all our experiments. 
For state-of-the-art baseline LVNet, we utilize code from \url{https://github.com/jongwoopark7978/LVNet}.
We discuss all steps in our proposed algorithms in detail while also releasing relevant code. All evaluations we perform are on public datasets accessible by all (some behind evaluation servers to prevent test set contamination). Our code is also publicly available at \url{https://github.com/kahnchana/mvu}.  

\section*{Contributions}
% \vspace{-0.2em}
KR led the project by building the preliminary ideas followed by performing most experiments and evaluations.
XL contributed to ideas on off-the-shelf model usage, setup all robotics evaluations, debugged many issues, and discussed all aspects of the project.  
KK contributed to ideas on VLM prompting and templating of textual data, setup several baseline replications, reviewed code, and discussed all aspects of the project.
MR organized the project, set the research direction, and discussed all aspects of the project idea, scope, and implementation.

\section*{Acknowledgments}
This work was supported by Electronics and Telecommunications Research Institute (ETRI) grants funded by the Korean government
[24ZR1100, A Study of Hyper-Connected Thinking Internet Technology by autonomous connecting, controlling and evolving ways]. This work was also supported by the Institute of Information \& Communications Technology Planning \& Evaluation (IITP) grant funded by the Korea government(MSIT) (No. RS-2024-00336738, Development of Complex Task Planning Technologies for Autonomous Agents).

We thank all members of the Robot Learning Lab at Stony Brook University for support, feedback and guidance. 
In particular, we would like to thank Jongwoo Park for technical feedback and support. We also thank Si Chen for logistical help and encouragement throughout the project.

\bibliography{main}
\bibliographystyle{iclr2025_conference}

\newpage
\appendix
\noindent {\huge \textbf{Appendix}}

\renewcommand{\thetable}{A.\arabic{table}}
\renewcommand{\thefigure}{A.\arabic{figure}}
\setcounter{table}{0}
\setcounter{figure}{0}
\setcounter{page}{1}



\section{Prompting and Template Operations}
\label{app:template}
In \Cref{subsec:obj_modal} and \Cref{subsec:oc-modalities}, we utilize 3 distinct prompts and fusion templates for generating joint textual inputs to be processed by the LLM. The 3 distinct prompt categories correspond to our Global Object Information~($x_{\mathrm{GOI}}$), Object Spatial Location~($x_{\mathrm{OSL}}$), and Object Motion Trajectory~($x_{\mathrm{OMT}}$) modalities. We first describe our exact templates as Python pseudo-code in \Cref{tab:modal_template}.

\begin{table}[ht]
\centering
\small
    \centering
    \begin{tcolorbox} 
        \centering
        % \small
        \hspace{-6mm}
        \begin{tabular}{p{0.99\columnwidth}}
        \textbf{Global Object Information~($x_{\mathrm{GOI}}$)} \\
        \texttt{\magenta{"Consider following objects in video to answer the question:"} + $\backslash$ }\\
        \texttt{\magenta{", "}.join(\blue{GOI\_data}) + \magenta{". "} + \blue{\texttt{task\_question}}}\\
        \rule[0.25\baselineskip]{\textwidth}{1pt}
        \textbf{Object Spatial Location~($x_{\mathrm{OSL}}$)} \\
        \texttt{\magenta{"Consider following objects with spatial location as }}\\
        \texttt{\magenta{(x, y, area) coordinates in video to answer the question:"} + $\backslash$ }\\ 
        \texttt{\magenta{", "}.join(\blue{OSL\_data}) + \magenta{". "} + \blue{\texttt{task\_question}}} \\
        \rule[0.25\baselineskip]{\textwidth}{1pt}
        \textbf{Object Motion Trajectory~($x_{\mathrm{OMT}}$)} \\
        \texttt{\magenta{"Consider following objects moving along (x, y, area) trajectories}}\\
        \texttt{\magenta{in video to answer the question:"} + $\backslash$ }\\
        \texttt{\magenta{", "}.join(\blue{\texttt{OMT\_data}}) + \magenta{". "} + \blue{\texttt{task\_question}}} \\
        \end{tabular}
    \end{tcolorbox}
    \vspace{-2mm}
    \caption{\textbf{Prompt templates for three textual modalities.}}
    \label{tab:modal_template}
\end{table}

The above templates depend on each of the modalities represented in textual form (i.e. \blue{\texttt{GOI\_data}}, \blue{\texttt{OSL\_data}}, \blue{\texttt{OMT\_data}}). We describe their exact textual forms next using examples in \Cref{tab:modal_example}.

\begin{table}[ht]
\centering
    \centering
    \small
    \begin{tcolorbox} 
        \centering
        % \small
        \hspace{-6mm}
        \begin{tabular}{p{0.99\columnwidth}}
        \texttt{\blue{GOI\_data} = [\magenta{"person"}, \magenta{"oven"}, \magenta{"dishwasher"}, \magenta{"sink"}, \magenta{"countertop"}, 
            \magenta{"dish"}, \magenta{"box"}, \magenta{"scissors"}, \magenta{"drain"}, \magenta{"hand"}, \magenta{"stove"}]}\\
        \rule[0.25\baselineskip]{\textwidth}{1pt}
        \texttt{\blue{OSL\_data} = [\magenta{"stove located at (0.52, 0.64, 0.595)"},} \\ 
\texttt{\qquad\qquad\qquad \magenta{"sink located at (0.56, 0.64, 0.211)"},}\\ 
\texttt{\qquad\qquad\qquad    \magenta{"countertop located at (0.63, 0.79, 0.308)"},}\\
\texttt{\qquad\qquad\qquad    \magenta{"box located at (0.46, 0.65, 0.142)"},}\\ 
\texttt{\qquad\qquad\qquad    \magenta{"dishwasher located at (0.5, 0.5, 0.991)"},}\\ 
\texttt{\qquad\qquad\qquad    \magenta{"dish located at (0.41, 0.75, 0.077)"},}\\ 
\texttt{\qquad\qquad\qquad    \magenta{"person located at (0.47, 0.76, 0.282)"} ]}\\
        \rule[0.25\baselineskip]{\textwidth}{1pt}
        \texttt{\blue{OMT\_data} = [\magenta{"stove trajectory: (0.5,0.5,0.991)->(0.51,0.69,0.397)}}  \\
        \texttt{\magenta{\qquad\qquad\qquad ->(0.54,0.73,0.396)"}, }  \\
        \texttt{\magenta{\qquad\qquad\qquad dish trajectory: (0.55,0.62,0.096)->(0.11,0.65,0.079)"}, }\\
\texttt{\qquad\qquad\qquad\qquad\qquad\qquad\qquad\qquad\qquad .}\\
\texttt{\qquad\qquad\qquad\qquad\qquad\qquad\qquad\qquad\qquad .}\\
\texttt{\qquad\qquad\qquad\qquad\qquad\qquad\qquad\qquad\qquad .}\\
        \texttt{\magenta{\qquad\qquad\qquad"dish trajectory: (0.41, 0.75, 0.077)"},}\\
        \texttt{\magenta{\qquad\qquad\qquad"person trajectory: (0.54,0.81,0.34)->(0.49,0.72,0.339)}}\\
\texttt{\qquad\qquad\qquad\magenta{           ->(0.54,0.84,0.157)->(0.23,0.71,0.176)}}\\
\texttt{\qquad\qquad\qquad\magenta{           ->(0.51,0.79,0.232)->(0.52,0.78,0.266)}}\\
\texttt{\qquad\qquad\qquad\magenta{          ->(0.39,0.64,0.558)->(0.54,0.82,0.184)"}]}\\
        \end{tabular}
    \end{tcolorbox}
    \vspace{-2mm}
    \caption{\textbf{Prompt examples for three textual modalities.}}
    \label{tab:modal_example}
\end{table}

In this example (for a single video), the \texttt{GOI\_data} list contains 11 distinct object categories discovered across all 8 selected frames for this video. In \texttt{OSL\_data}, this category list is grounded to each frame using our object detector. We apply this on 16 uniformly sampled frames as opposed to only 8 used with the captioner. While this stage removes some categories (which we assume could be object hallucinations \citep{RanLearningtoLoc23}), it also identifies categories at the instance level. We draw attention to the two different instances of a dish in our \blue{\texttt{OSL\_data}} for this example. Also, note that the single spatial coordinate reflects the average location of that object across all 16 (or the number of frames it is detected in) following the setup in \cite{RanLearningtoLoc23}. Our object tracks calculated across frames are utilized for this averaging (i.e. distinguish the two dishes).  
For our \blue{\texttt{OMT\_data}}, we again leverage our tracks where each object instance is matched across frames and construct explicit sequences of object locations across frames. While ignoring the actual frame indexes, we only consider the object trajectory using frames where they are detected. Note that an object trajectory could be limited to a single location or a variable number of multiple locations. Also, for these trajectories, we introduce an additional scale factor for each object location. This scale factor is the ratio of the object bounding box area to image area, i.e. $(\texttt{obj\_width} * \texttt{obj\_height}) \div \texttt{im\_size}$. This is introduced with an aim of possibly providing some level of depth information. In terms of generating object tracks, we utilize intermediate features from our object detector and perform feature matching based tracking.   

\section{Details on Pretrained Models and Datasets}
\label{app:models_data}
We describe in detail the pretrained models used to construct our framework as well as the multiple datasets used in evaluating our framework. 

\noindent \textbf{Models:}
Our framework utilizes three distinct off-the-shelf models for its various operations, namely \textit{a}) an LLM / VLM for likelihood selection, \textit{b}) a generative VLM for extracting object list from a frame, and \textit{c}) an open-vocabulary detector for object localization. We use \texttt{LLaVA-v1.5-13B} \citep{liu2023llava} for likelihood selection and frame object list generation. For object localization, we use \texttt{OWL-ViT-B/32} \citep{Minderer2022SimpleOO}. Unless explicitly specified, we use the above setup in all our experiments. 
Variants of our framework uses LLMs \texttt{Llama-2-7b-Chat}, \texttt{Gemma-7b-IT}, and \texttt{Mistral-7B-Instruct} (default) for likelihood selection. 
Apart from these off-the-shelf models, our framework involves zero additional training. We also reiterate that no components of our framework undergo any form of video-level training.

\vspace{0.5em}
\noindent \textbf{Datasets:}
We use multiple video datasets for evaluation under question-answering or n-way classification settings.
For video question answering, we select two datasets focused on long-form videos: EgoSchema \citep{Mangalam2023EgoSchemaAD}, NExT-QA \citep{dataset_xiao2021nextqa}. EgoSchema is a long-form ego-centric video question-answering benchmark, consisting of a 500-video public subset (EgoSchema-S) and a full 5000+ video evaluation set (EgoSchema-F) accessed only through evaluation servers. This dataset spans over 250 hours and is specially constructed to ensure that \textit{questions require awareness of a longer temporal window for correctly answering} \citep{Mangalam2023EgoSchemaAD}. Example images of EgoSchema are shown in \Cref{fig:dt_example}.
NExT-QA similarly contains long-form videos with a focus on requiring causal \& temporal action reasoning as well as common scene comprehension for correctly answering. It contains a validation set (NExT-QA-V) of 4996 video-questions pairs and a test set (NExT-QA-T) of 8564 video-question pairs.
We also use a series of robotics datasets from the Open X-Embodiment robotics dataset \citep{open_x_embodiment_rt_x_2023} for video question answering in a different domain (more detail in \Cref{subsec:robotics}). 
In only one of our ablations aimed at analyzing the motion understanding aspect of our framework, we utilize a fine-grained action recognition dataset, Something-Something-v2 \citep{Goyal2017TheS}, that contains 174 action categories focusing on object motions by replacing object category nouns with `\textit{something}' in textual descriptions of each action category.

\section{Details on Baselines}
\label{app:baselines}
In \Cref{subsec:eval_longvid}, we evaluate performance on long-video understanding tasks using work in \cite{zhang2023llovi} and \cite{wang2023vamos} as two baselines for comparison. However, both these methods utilize closed-source, proprietary LLMs (i.e. GPT-4) with parameter counts on the order of trillions (over 100X our model size) deeming their direct comparisons unfair. In the case of \cite{zhang2023llovi}, we replicate their method (using their open-source repository and pre-trained models following \cite{Kahatapitiya2024}) utilizing an open-source LLM of comparable parameter count as our framework. For \cite{wang2023vamos}, the authors directly report results for a variant with a similar parameter count as ours. We utilize these evaluations as our point of comparison. 

We also replicate prior work, LVNet \citep{Park2024TooMF}, that exhibits state-of-the-art results. For this, we use their official code (\url{https://github.com/jongwoopark7978/LVNet}) and integrate our \modelname framework over this baseline.

We highlight that re-implementations of these baselines utilize common LLMs / VLMs as our MVU framework followed by identical evaluation protocols to ensure fair comparison. 

\section{Robotics Domain Dataset Details}
\label{app:openx}
The Open X-Embodiment dataset is an extensive collection of visuomotor robotics datasets, encompassing a wide range of tasks and environments. 
It is designed to facilitate research in visuomotor control, providing rich sensory inputs and motor outputs for training and testing robotic systems. 
However, the videos are usually taken in a controlled environment and they do not always contain meaningful objects, which makes the samples in the dataset usually out of general video distribution (See~\Cref{fig:dt_example}).

For our analysis, we specifically select datasets within this collection that contain expert episodes accompanied by corresponding language instructions and adapt them into video classification datasets. 
We treat each trajectory within the dataset as a video clip, with its associated language instruction serving as the video caption (classification label). 
For each episode, the model is tasked with identifying the correct language instruction from a set of five options, considering a video clip uniformly downsampled to 8 frames. 
The incorrect options are randomly chosen from the dataset to ensure a diverse and challenging selection.
In instances where the datasets have multiple cameras for each episode, we treat the videos captured by each camera as distinct datasets.

\section{Discussion on Modality Constrained Evaluation}
\label{app:modality}

We evaluate the two modality-constrained variants of our approach, \llmbaseline and \vlmbaseline (details in \Cref{subsec:modal_intro}) and summarize these findings in \Cref{tbl:baseline}. We uncover surprisingly strong performance of both variants on two long-video understanding benchmarks. Note how these approaches use no video-specific information to generate predictions. 

We highlight how our best \llmbaseline variant achieves performance significantly higher than random selection (+25.8\% on EgoSchema-S / +20.1\% on NextQA-T) using zero visual information. This indicates the large portion of questions in existing video-QnA benchmarks that can be answered correctly purely using world knowledge. 
We also highlight our single frame variant performing on par with some existing state-of-the-art (gray). In particular, for EgoSchema-S we outperform \cite{zhang2023llovi} which uses information extracted from 180 frames per video incurring an inference cost over 100 times higher than ours. In light of these findings, we argue that long video understanding approaches in particular must focus on learning information beyond what a single frame baseline can achieve. 

We also evaluate these same modality-constrained variants on robotics domains tasks and report these results in \Cref{tbl:app_robotics}. In contrast to the results on standard long-video QnA benchmarks, the robotics domains results are more aligned with intuition: the no-visual input \llmbaseline performs on par with random and the \vlmbaseline marginally outperforms random selection. 

We attribute this difference in performance to the nature of robotics domain tasks. They tend to involve controlled environments with often naive, meaningless tasks purely for robot testing purposes. These may not necessarily align with human commonsense or other constraints dependent on knowledge of our world. Therein, the clear ability of LLMs to solve general everyday video tasks (e.g. EgoSchema, NextQA performance in \Cref{tbl:baseline}) using its world knowledge may not be applicable to robotics domain tasks. 
Utilizing different domain benchmarks, in particular robotics tasks, provides a much more representative evaluation of LLM based video QnA approaches.

\begin{table}[t]
\centering
\small
\begin{minipage}{0.48\textwidth}
\vspace{0.5em}
\caption{\textbf{Modality Constrained Variants on Robotics Domain:} 
We evaluate our modality constrained baselines on the robotics domain tasks and report accuracy (\%). Note that a weighted sum over multiple tasks is reported here (similar to \Cref{tbl:robotics}). Note the minimal increase over random for the variants in contrast to generic video benchmarks.
}
\label{tbl:app_robotics}
\end{minipage}
\hspace{0.01\textwidth}
\begin{minipage}{0.48\textwidth}
\vspace{-0.5em}
\def\arraystretch{1.2}  % height
\setlength\tabcolsep{0.9em}  % width
\scalebox{0.95}{
\begin{tabular}{lccc}
\toprule
Method        & Visual & Frames & Accuracy \\ \midrule
Random        & -      & -      &  22.1  \\
\llmbaseline  & \xmark & -      &  21.9  \\
SF-VLM        & \cmark & 1      &  23.5  \\
MVU           & \cmark & 16     &  30.4  \\ \bottomrule
\end{tabular}
}
\end{minipage}
\vspace{1.0em}
\end{table}

\section{Likelihood Selection}
\label{app:ls}
In this section, we present the prompts and templates used to adapt likelihood selection inference \citep{Robinson2022LeveragingLL} to our video QnA tasks. Our experimentation shows significantly higher sensitivity (to prompt style) of LLM performance on QnA tasks when using like likelihood selection in comparison to sequential text generation (consistent with findings in \cite{Robinson2022LeveragingLL}). We evaluate a series of different prompt templates on the EgoSchema and Next-QA dataset to discover optimal combinations. The best prompt templates used in our final framework are presented in \Cref{app:ls_prompts} as Python pseudo-code. For Next-QA in particular, the average zero-shot accuracy could vary from 35\% to 55\% with slight variations of the prompt templates. 

Our optimal prompt templates for the standard video QnA tasks also generalized to our robotics domain QnA tasks. Nevertheless, we highlight the possibility of needing some prompt template tuning when applying our framework to different domains. We also note that while our prompt selection process was guided by heuristics and experimentation, there may be other similar prompts performing equally well or surpassing our optimal selection.

\subsection{Implementation Details}
We revisit the generation process of autoregressive LLMs and their visual extensions (VLMs). 
They commonly use iterative prediction of next tokens conditioned on prior outputs to generate complete natural language outputs. Such a generation process is usually modeled as sampling from a conditional likelihood shown as \Cref{eq:ntp}, where $\hat{y}^j$ stands for the $j^{\mathrm{th}}$ token in a textual sequence $\hat{y}$ autoregressively generated by the model.
\begin{align}
    \label{eq:ntp}
    P(\hat{y}|x_t) &= \prod_j P(\hat{y}^j|\hat{y}^{1, ..., j - 1}, x_t)
\end{align}
The dependency on prior output $\hat{y}^{1, ..., j - 1}$ makes this process both computationally costly and redundant in the case of choice-based answer selection tasks. Alternately, given the closed set of $Y$ in choice-based selections tasks, we formulate $P(y_i|x_t)$ for any $y_i \in Y$ with no dependency on any model generated output ($\hat{y}$) as, % prior model outputs ($\hat{y}$) as, 
\begin{align}
    \label{eq:cl}
    P(y_i|x_t) &= \prod_j P(y^j_i|y^{1, ..., j - 1}_i, x_t)
\end{align}
Assume a perfect LLM, intuitively when $y_i$ is a proper answer to the question $x_t$ (say $y_i=y_g$), the conditional likelihood $P(y_i|x_t)$ should be larger than any other $P(y_w|x_t)$ where $y_w$ is a wrong answer to question $x_t$.
In fact, modern LLMs are trained with a similar objective \citep{Radford2018ImprovingLU}.
Motivated by this standard LLM training objective, we estimate the relative numerical scales of conditional likelihood on different answers $P(y_i|x_t)$ using a cross-entropy error $e_i$, given their equivalence (negative log-likelihood and multiclass cross-entropy, see Section 4.3.4 in \cite{Bishop2006PatternRA}).
We calculate $e_i$ with a single forward pass of LLM without detokenization and the selection can be made by simply picking up the answer with the lowest error, equivalent to the highest conditional likelihood among all the answers. 

\begin{table}[t]
\centering
    \centering
    \begin{tcolorbox} 
        \centering
        % \small
        % \hspace{}
        \begin{tabular}{p{0.99\textwidth}}
        \\
        \texttt{\blue{prompt\_list = $\backslash$}} \\
        \texttt{\blue{\qquad[f"Response \{idx\}:\{val\}" for idx, val in enumerate(prompt\_list)]}} \\
        \\
        \texttt{\blue{system\_prompt = $\backslash$}} \\
        \texttt{\blue{\qquad "Considering given frames of a long video, select the most}} \\
        \texttt{\blue{\qquad \ suitable response to the following question from the five}} \\
        \texttt{\blue{\qquad \ options provided."}} \\
        \\
        \texttt{\blue{response\_template = $\backslash$}} \\
        \texttt{\blue{\qquad "The correct response best answering the question about the given}} \\
        \texttt{\blue{\qquad \  video is "}} \\
        \\
        \texttt{\blue{task\_prompt = "Question: \{qs\}" + ''.join(prompt\_list)}} \\
        \\
        \texttt{\blue{qs = system\_prompt + task\_prompt + response\_template}} \\
        \end{tabular}
    \end{tcolorbox}
    \caption{
    \textbf{Likelihood Selection Sample Prompt Templates.}
    Variables \textit{qs} and \textit{prompt\_list} refer to per sample question and choice list respectively. 
    }
    \label{app:ls_prompts}
\end{table}

This sets the ground for \textit{Likelihood Selection}, also referred to as Cloze Promting in \cite{Robinson2022LeveragingLL}, first illustrated with a toy example in \Cref{fig:selection}, where the task is vanilla question-answering with only textual context and the model takes one question $x_t$ as well as $M=5$ candidate answers $y_{1, ..., 5}$. To find the best answer, we simply concatenate the question with each candidate independently ($s_i = \mathrm{concat}\left(x_t, y_i\right)$ ) and pad them into a batch $\{s_{1, ..., 5} \}$. 
Then the LLM takes the batch of five samples with causal attention masks and performs one inference forward pass, resulting in five shifted logits $\{p_{1, ..., 5} \}$. 
Next, we shift the input sequence $s_i$ to align the logits $p_i$ and calculate the average cross-entropy error only on tokens of $y_i$ Finally, the answer with the smallest $e_i$ will be picked up as the best answer.
The method can be formulated as in \Cref{eq:ls} using equivalence of negative log-likelihood to cross-entropy in \Cref{eq:nll}. Here $n_i$ stands for the token sequence length of $y_i$ and $p_i^j$ stands for logits of the $j^{\mathrm{th}}$ token in $p_i = V(\mathrm{concat}\left(x_t, y_i\right))$ with logits limited to only those of $y_i$. 
\begin{align}
% \text{log } P(y_i|x_t) &= \sum_j P(y^j_i|y^{1, ..., j - 1}_i, x_t) \\
% e_i(y_i) &= \mathrm{CrossEntropy}(p_i, y_i) = \frac{1}{n_i} \sum_{j}^{n_i} \left(\mathrm{CrossEntropy}(p_i^j, y_i^j) \right) \\
\label{eq:nll}
e_i(y_i) = \mathrm{CE}(p_i, y_i) &= \frac{1}{n_i} \sum_{j}^{n_i} \left(\mathrm{CE}(p_i^j, y_i^j) \right) 
\approx \sum_{j}^{n_i} - \mathrm{log} \: P(y_i|x_t) \\
\label{eq:ls}
\mathcal{F}_{\text{LS}}(Y, x_t) &= \argmax_{y_i \in Y} P(y_i|x_t) = \argmin_{y_i \in Y} e_i(y_i)
\end{align}
 % \xl{please revise, I can not come up with any better words at this time!!!}
In summary, Likelihood Selection performs one single forward pass to extract the network logit outputs, calculates error ($e_i$) on each choice,
% to a training step but without calculating gradients, back-propagation, or changing any parameters,
and selects the choice with the lowest error. Note that our method does not utilize any iterative autoregressive generation using the LLM. 
This results in considerable speed-ups for inference time. 
We also obtain the additional advantages of avoiding LLM hallucinations and deviations from expected output formats over iterative generation strategies applied to similar visual tasks \citep{Hanu2023LanguageAT} leading to better accuracy (see Tab. 6.).
In \Cref{subsec:modal_intro}, we demonstrate both our speed gains and performance improvements. 

Furthermore, Likelihood Selection is a generic method that can be easily extended to autoregressive VLMs, and in principle, there is no reason it could not also be used with extra modalities besides language. We validate this across all our experiments using the multimodal \modelname framework. 

\subsection{Distinction from Exact Match}

As described in the previous section, likelihood selection uses a likelihood measure which is the likelihood (probability) of the model generating the given sentence (as opposed to being an exact match). 
This likelihood measure is also used as the training loss when training LLMs. Given how LLMs trained with this loss (i.e. all decoder based LLMs such as LLaMA, Gemini, GPT) are highly effective at handling semantic meaning, it follows that this loss can capture semantic meaning.
This likelihood measure is calculated within the LLM latent space. This is equivalent to the probability (or likelihood) of that answer being generated by the LLM conditioned on the input question. We derive this in detail in Appendix F. Relating to the same example, this means that likelihood is an estimate of how likely the model would predict ‘C is washing plates’ as opposed to making that exact match. This means predictions closer to the target such as ‘C is cleaning dishes’ would also gain high likelihood values.

In fact, we validate this second point through a toy example. We provide an LLM with the question "X is cleaning dishes in the kitchen. What is X doing? a) washing plates, b) cleaning laundry, c) painting dishes. The correct choice is:" and calculate the likelihood for each of the 3 responses. The calculated likelihoods are 0.996, 0.006, 0.007 for a, b, c respectively (highest is selected), despite response (a) having no common words with the original statement unlike (b) and (c). This illustrates the ability of likelihood selection to capture semantic meanings.

\subsection{Detailed Prompting Example}

We also note that while different choices are repeated along the batch, our likelihood implementation actually follows prior approaches where all answer candidates are fed together to the language model in addition to organizing the Q-A pairs in a batch dimension. Taking one simplified toy example, given a question “Where is the dog?” and answers “mat, table, bench”, we use three queries along batch dimension as:
\begin{itemize}
    \item \texttt{Where is the dog? Select the correct response from: a) mat, b) table, c) bench. The correct response is a) mat.}
    \item \texttt{Where is the dog? Select the correct response from: a) mat, b) table, c) bench. The correct response is b) table.}
    \item \texttt{Where is the dog? Select the correct response from: a) mat, b) table, c) bench. The correct response is c) bench.}
\end{itemize}
In fact, applying likelihood selection without such prompting leads to significantly low performance for some datasets. We show this in \Cref{ablate:ls} which we repeat here as \Cref{app:repeat_tbl_6}.

\begin{table}[t]
\centering
\small
\caption{
\textbf{Ablating Answer Candidates in Prompt:}
We illustrate the importance of appropriate prompting when combining with likelihood selection, specifically for long video QnA tasks. Top-1 accuracy (\%) is reported on EgoSchema subset (ES-S) and NextQA test set (NQA-T).
}
\label{app:repeat_tbl_6}
\begin{tabular}{l|c|c}
\toprule
Dataset                          & ES-S & NQA-T \\ \midrule
No answer candidates in prompt   & 58.2 & 35.8  \\
With answer candidates in prompt & 60.3 & 55.4  \\ \bottomrule
\end{tabular}
\end{table}

\begin{table}[t]
\centering
\small
\caption{\textbf{Open-Ended Video QnA Evaluation}: 
We present results on the ActivityNet dataset \citep{Yu2019ActivityNetQAAD} that demonstrate strong performance of our proposed \modelname framework. Accuracy (\%) is reported. VT stands for video level training. We highlight how our \modelname framework utilizes no video level training for any of its components and surpassed multiple approaches that rely on video-language training. 
} 
\label{tbl:act_res}
\vspace{-0.5em}
\def\arraystretch{1.1}  % height
\setlength\tabcolsep{1.3em}  % width
\scalebox{0.90}{
\begin{tabular}{lccc}
\toprule
Method          & Zero-Shot & VT & ActivityNet-QA  \\ \midrule
JustAsk \citep{yang2021justask}               & \xmark & \cmark & 38.9 \\
FrozenBiLM \citep{yang2022frozenblim}         & \xmark & \cmark & 43.2 \\
VideoCoCa \citep{Yan2022VideoCoCaVM}          & \xmark & \cmark & 56.1 \\ \midrule
FrozenBiLM \citep{yang2022frozenblim}         & \cmark & \cmark & 24.7 \\
Video Chat \citep{2023videochat}              & \cmark & \cmark & 26.5 \\
LLaMA Adapter \citep{Zhang2023LLaMAAdapterEF} & \cmark & \cmark & 34.2 \\
Video LLaMA \citep{Zhang2023VideoLLaMAAI}     & \cmark & \cmark & 12.4 \\
Video-ChatGPT \citep{Maaz2023VideoChatGPTTD}  & \cmark & \cmark & 35.2 \\ 
LocVLM \citep{RanLearningtoLoc23}             & \cmark & \cmark & 37.4 \\ 
Video-LLaVA \citep{Lin2023VideoLLaVALU}       & \cmark & \cmark & 37.4 \\ 
VISTA-LLaMA \citep{Ma2023VistaLLaMARV}        & \cmark & \cmark & 37.4 \\ 
VideoChat-2 \citep{Li2023MVBenchAC}           & \cmark & \cmark & 37.4 \\ 
LLaMa-VID \citep{Li2023LLaMAVIDAI}            & \cmark & \cmark & 37.4 \\ 
LLoVi \citep{zhang2023llovi}                  & \cmark & \xmark & 41.8 \\  \rowcolor{Gray}
\modelname (ours)                             & \cmark & \xmark & \textbf{42.2}    \\ \bottomrule
\end{tabular}
}
% \vspace{-0.8em}
% \vspace{-1em}
\end{table}

\section{Open-Ended Video Question Answering}
\label{app:open_qa}

In this section, we explore the ability of our proposed \modelname framework to operate on open-ended video question answering (QnA) tasks. For this purpose, we evaluate on the Activity-Net dataset \citep{Yu2019ActivityNetQAAD} reporting the accuracy metric. We follow evaluation settings identical to \cite{Maaz2023VideoChatGPTTD} for these evaluations. 

Given the nature of open-ended QnA tasks (i.e. no answer choices, generate free form answers), we use standard generation instead of likelihood selection. We match the generated answers against ground-truth following \citep{Maaz2023VideoChatGPTTD}. We present these results in \Cref{tbl:act_res} where our \modelname achieves strong results and clear improvements over the similar LLM based approach from \cite{zhang2023llovi}. We compare against multiple recent approaches that use similar capacity LLMs \ VLMs for open-ended video QnA. 
We take these results as another indication to the generality of our \modelname framework on video QnA tasks beyond MCQ style. 

\section{Longer Video Question Answering}
\label{app:longvideobench}

While established long video benchmarks used as the key evaluations in numerous prior work \citep{wang2025videoagent,wang2024videotree,min2024morevqa,Park2024TooMF,zhang2023llovi,Kahatapitiya2024,Wang2023LifelongMemoryLL} limit to roughly 1-3 minute long videos, some newer datasets include even longer videos \citep{wu2024longvideobench}. We explore such even longer videos by evaluating our method on the LongVideoBench dataset \citep{wu2024longvideobench}. 

We select Phi-3-Vision-Instruct \citep{Abdin2024Phi3TR} as our baseline since it is the best performing model we can replicate within our compute budget. We note that larger sized models using significantly larger context lengths are difficult to replicate within academic compute restraints. Results using this baseline from \cite{Abdin2024Phi3TR} and our MVU framework integrated over it are presented in \Cref{app:tbl_lvb}. MVU gains clear performance gains in this longer video dataset.

\begin{table}[t]
\small
\centering
\begin{minipage}{0.40\textwidth}
\caption{
\textbf{LongVideoBench Evaluation:}
We integrate MVU with the baseline from \cite{Abdin2024Phi3TR} and highlight the additional performance improvements achieved by our MVU framework.
}
\label{app:tbl_lvb}
\end{minipage}
\hspace{0.01\textwidth}
\begin{minipage}{0.56\textwidth}
\centering
\begin{tabular}{l|c}
\toprule
Method                      & Acc (\%) \\ \midrule
Phi-3-Vision-Instruct \citep{Abdin2024Phi3TR}      & 49.7     \\
Phi-3-Vision-Instruct + MVU & 50.4     \\ \bottomrule
\end{tabular}
\end{minipage}
\end{table}

\begin{table}[t]
\centering
\small
\caption{\textbf{Ablation on Object Motion Trajectory (OMT) modality:} 
We perform this ablation on a different dataset given the motion focused aspect we explore. Accuracy (\%) reported on the motion-based SSv2 dataset clearly indicate the usefulness of the OMT modality in our \modelname framework.}
\label{ablate:motion}
% \vspace{-1.0em}
\def\arraystretch{1.2}  % height
\setlength\tabcolsep{1.1em}  % width
\scalebox{0.90}{
\begin{tabular}{lccc}
\toprule
Method                          & OMT    & Accuracy  \\ \midrule
Random                          &    -   & 0.6 \\ 
CLIP \citep{radford2021clip}     &    -   & 4.0 \\ 
MAXI \citep{lin2023match}        &    -   & 6.4 \\ \midrule
\modelname (ours)               & \xmark & 3.6 \\ \rowcolor{Gray}
\modelname (ours)               & \cmark & \textbf{7.2} \\ \bottomrule
\end{tabular}
}
\end{table}

\section{Additional Ablations}
\label{app:ablate_more}

In this section, we repeat part of our ablation from \Cref{ablate:modal} focused on the object motion trajectory modality inputs. We note that common video QnA benchmarks require minimal understanding of object motion to answer most questions. Our goal is to explore the value of motion information in a more relevant tasks.

Therein we investigate a new motion focused dataset, Something-Something-v2 \citep{Goyal2017TheS} (SSv2), only for this single ablation. The SSv2 dataset focuses on motion-based category discrimination, providing an ideal evaluation to measure the usefulness of our object motion trajectory modality. We benchmark on a subset of this dataset following \citep{lin2023match} and report these results in \Cref{ablate:motion}. Our results while exceeding their performance also indicate the clear performance gains obtained when injecting the object motion trajectory modality into our \modelname framework.

We also provide an ablation on frames used with our MVU framework in \Cref{app:tbl_frames}. Increasing the number of frames leads to improved performance in contrast to some prior works \citep{Mangalam2023EgoSchemaAD} highlighting how our information fusion pipeline allows better utilization of the LLM context length. Additionally, the lightweight object detector and tracker used in MVU allows scaling the number of frames with a lesser increase in inference time.

\begin{table}[t]
\centering
\small
\caption{
\textbf{Frame Count Ablation:}
We illustrate the importance of appropriate prompting when combining with likelihood selection, specifically for long video QnA tasks. Top-1 accuracy (\%) is reported on EgoSchema subset (ES-S) and NextQA test set (NQA-T).
}
\label{app:tbl_frames}
\begin{tabular}{l|c|c|c}
\toprule
Method & Frames & EgoSchema-S & Time (s) \\ \midrule
MVU    & 16     & 60.3        & 2.42     \\
MVU    & 32     & 60.4        & 2.48     \\
MVU    & 64     & 60.4        & 2.60     \\ \rowcolor{Gray}
MVU    & 128    & 61.2        & 2.81     \\ \bottomrule
\end{tabular}
\end{table}

\section{Tokenization in LLMs}
\label{app:tokenization}
Most modern LLMs utilize Byte-Pair Encoding (BPE) tokenization \citep{Sennrich2015NeuralMT} to convert natural language into discrete tokens that are mapped to vocabulary embeddings. This process is learned from language itself and the resulting tokenization may sometimes break complete words into pieces (e.g. \texttt{example $\rightarrow$ ex-am-ple}). Given our utilization of logits extracted from an LLM forward pass, we note that each logit corresponds to a single token, which may at times be the embedding of some meaningless word piece. However, our calculation of a joint likelihood across a sequence of tokens ensures a meaningful process, which is validated by our strong results.

\section{LLM Context Length}
\label{app:llm_context}

Using LLMs for long video understanding has proven successful \citep{zhang2023llovi,wang2024videotree} but handling long context lengths remains a key issue \citep{Mangalam2023EgoSchemaAD,Park2024TooMF}, often leading to lower performance when additional frame information is provided to the LLM. This draws importance to frame selection, but we argue that alternate forms of information bottlenecks can also provide improvements, often complementary to frame selection.  

In MVU, instead of naively collecting all information within a frame, we only collect object centric spatial and motion information, allowing to process more frames at a fixed context length. In other words, MVU information extraction from multiple frames can be viewed as an alternative to frame selection. This is because our object information extraction indirectly acts as an information bottleneck similar to frame selection. For frames without objects of interest, no information is extracted. For multiple frames containing the same object (identified by our object tracker), the repetitive information is removed. This resembles the idea of selecting useful information from multiple frames. 

In fact, when comparing the average token length for a similarly performing baseline (implemented under identical settings using a common LLM), we use less tokens (context length) to achieve similar results. We show these results in \Cref{app:tbl_context_length}.

\begin{table}[t]
\small
\centering
\begin{minipage}{0.48\textwidth}
\caption{
\textbf{Context Length Comparison:}
We compare the context length used (i.e. number of tokens) to achieve similar results with LLoVi \citep{zhang2023llovi} as opposed to our MVU. We achieve better performance utilizing less tokens.
}
\label{app:tbl_context_length}
\end{minipage}
\hspace{0.01\textwidth}
\begin{minipage}{0.48\textwidth}
\centering
\begin{tabular}{l|c|c}
\toprule
Method & Average Tokens & ES-F (\%) \\ \midrule
LLoVI  & 1940           & 33.5     \\
MVU    & 1124           & 37.6     \\ \bottomrule
\end{tabular}
\end{minipage}
\end{table}

\section{Additional Baselines}

We implement a multi-frame baseline directly using LLaVA-1.5 \citep{liu2023llava} with no video level training. These results are reported in \Cref{app:tbl_multi_llava}. Results indicate that directly adding multiple frames to a VLM with no video level training does not lead to improved performance.
Similar trends are observed in prior work \citep{Kahatapitiya2024}.
These findings highlight the importance of careful per-frame information extraction and cross frame information fusion proposed in our MVU.

\begin{table}[t]
\small
\centering
\begin{minipage}{0.48\textwidth}
\caption{
\textbf{Multi-Frame LLaVA Baseline:}
We implemented multi-frame variants of LLAVA \citep{liu2023llava} with no video level training. Results indicate that without any video level training such naive extension does not lead to results improvements. 
}
\label{app:tbl_multi_llava}
\end{minipage}
\hspace{0.01\textwidth}
\begin{minipage}{0.48\textwidth}
\centering
\begin{tabular}{l|c|c}
\toprule
Method & Frames & ES-S \\ \midrule
LLaVA  & 1      & 55.8 \\
LLaVA  & 8      & 53.4 \\
LLaVA  & 16     & 46.2 \\
LLaVA  & 32     & 40.2 \\ \bottomrule
\end{tabular}
\end{minipage}
\end{table}

\section{Extension to Robot Control}
\label{app:mvur}

\begin{figure}[h]
    \centering
    \includegraphics[width=\linewidth]{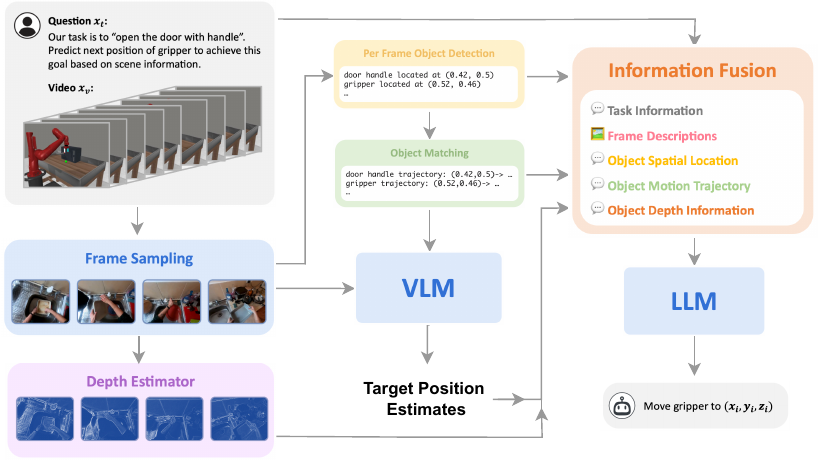}
    \vspace{-1.5em}
    \caption{\textbf{MVU Robot Control Extension:} 
    We adapt MVU for robot manipulation (MVU-R) by framing control as a video question answering task, enabling zero-shot action prediction via vision-language prompting.
    }
    \label{app:robot-arch}
    \vspace{0.5em}
\end{figure}

We extend our MVU framework to robot manipulation tasks, as illustrated in Figure~\ref{app:robot-arch}. Inspired by prior work such as LLaRA \citep{Li2024LLaRASR}, we formulate robot control—specifically, manipulation with a robot arm in tabletop environments—as a question-answering (QA) problem. This formulation allows us to directly leverage our MVU architecture, originally designed for video QA, to solve robotic manipulation tasks with minimal modification. 
We refer to this as MVU-R.

In contrast to the default MVU, our VLM here generates an estimate of the next gripper position, tagged \texttt{Target Position Estimate}. This is generated as an $(x,y)$ coordinate pair in image-space, motivated by prior work such as LLaRA \citep{li2024llara}. Paired with past object motion information and depth information, these coordinate sequences are processed by an LLM to provide a final gripper position. We provide a detailed example in \Cref{app_sec:robot_impl}. 

We conduct experiments in the MetaWorld~\citep{yu2019meta} simulated environment and evaluate on two manipulation tasks, following prior work such as AVDC~\citep{Ko2023LearningTA} and LTM~\citep{Rana2025LangToMo}. MetaWorld is a benchmark suite consisting of diverse robotic manipulation tasks performed by a simulated Sawyer robot arm in tabletop scenes. Each task provides access to egocentric RGB videos, object bounding boxes, and environment metadata.

To support 3D spatial understanding, we augment the input with depth maps from the simulator, similar to the AVDC baseline. In real-world applications, such depth information can be extracted using off-the-shelf depth estimation models~\citep{Rana2025LangToMo, Ko2023LearningTA}.

Table~\ref{app:robot-results} presents the results of our MVU-based robot control system. Remarkably, our model—without any robotics-specific fine-tuning—demonstrates promising zero-shot performance on these tasks, using only task-specific prompting. These findings suggest the potential of VLM/LLM-based models to serve as general-purpose, off-the-shelf agents for embodied control. We also emphasize how our MVU-R framework performs these tasks zero-shot while all prior works require some form of model training on expert task demonstrations (videos or action trajectories) from the environment.

\begin{table}[h]
\small
\centering
\caption{
\textbf{MVU Robot Control Extension:}
We compare task success rates on two MetaWorld tasks: \textit{Door-Open} and \textit{Door-Close}. These tasks are selected following work in AVDC~\citep{Ko2023LearningTA} and LTM~\citep{Rana2025LangToMo}. Our MVU extension (MVU-R) achieves competitive zero-shot performance without any robotics-specific training, using only vision-language prompting. Results for baseline methods are taken from prior work, AVDC~\citep{Ko2023LearningTA} and LTM~\citep{Rana2025LangToMo}. While MVU-R achieves competitive results under zero-shot settings in simple object movement tasks (open / close door), it fails in complex object manipulation tasks. Tasks that require no gripper movement are marked ``gripper \xmark''. 
}
\label{app:robot-results}
\def\arraystretch{1.2}  % height
\setlength\tabcolsep{1.2em}  % width
\begin{tabular}{lccccc}
\toprule
\multirow{3}{*}{Method} & \multirow{3}{*}{Zero-Shot} & \multicolumn{4}{c}{Task} \\ \cline{3-6} 
            &           &
            \begin{tabular}[c]{@{}l@{}}Door-Open\\ (gripper \xmark)\end{tabular}   &
            \begin{tabular}[c]{@{}l@{}}Door-Close\\ (gripper \xmark)\end{tabular}  &
            \begin{tabular}[c]{@{}l@{}}Basket-Ball\\ (gripper \cmark)\end{tabular} &
            \begin{tabular}[c]{@{}l@{}}Shelf-Place\\ (gripper \cmark)\end{tabular} 
            \\ \midrule
BC-Scratch  & \xmark  & 21.3\%      & 36.0\%   & 0.0\%   & 0.0\%  \\
BC-R3M      & \xmark  & 1.3\%       & 58.7\%   & 0.0\%   & 0.0\%  \\
UniPi       & \xmark  & 0.0\%       & 36.0\%   & 0.0\%   & 0.0\%  \\
AVDC        & \xmark  & 72.0\%      & 89.3\%   & 37.3\%  & 18.7\%   \\
LTM         & \xmark  & 76.0\%      & 94.7\%   & 38.0\%  & 15.2\%  \\  \rowcolor{Gray}
MVU (ours)  & \cmark  & 66.7\%      & 77.3\%   & 0.0\%   & 0.0\%  \\ \bottomrule
\end{tabular}
\end{table}

\subsection{Implementation Details}
\label{app_sec:robot_impl}
Our MVU-R framework sequentially processes frames from a robot environment. At each time step, all objects are detected in current frame, connected to tracks in past frames, and processed by a VLM (both image and object motion tracks) to estimate a next gripper position. The VLM is prompted \texttt{Based on provided motion information of all objects in scene, estimate next position of gripper. Only provide (x,y) coordinate.}. This is used along with the MVU default system prompts that describe the motion trajectory format and specify additional video relevant information (even for robotics, we consider the past frame sequence as a video). 

The next information fusion stage is similar to MVU, except for depth information and gripper position estimates. 
All object (x,y) coordinates are augmented with a depth $z$ coordinate obtained from a depth map of the image.
This VLM output (after suitable templating operations to extract only x,y) is added as a gripper next position estimate. 
This information is processed by our LLM, similar to MVU, except in MVU-R a gripper position as $(x,y,z)$ is generated instead of an answer to a question.

\subsection{Low Level Policy}
As described previoisly, our MVU-R framework provides 3D coordinates, with $(x,y)$ based on image space and $z$ based on image-based depth map. We utilize the hand-crafted mapping of 3D locations in image-space to MetaWorld action vectors from AVDC~\citep{Ko2023LearningTA} and LTM~\citep{Rana2025LangToMo} in our work. This also allows us fair and straightforward comparison to the evaluations from these prior works.

\subsection{Limitations in MVU Robot Control}
The key limitations of MVU-R are prompt sensitivity and lacking grasper awareness. 
When extending MVU-R to novel tasks, our system requires extensive hand-crafting of prompts. We consider this a trade-off for the zero-shot, training-free nature: our framework has never been trained on robotics tasks. 
MVU-R also currently cannot handle intelligent gripper manipulation (i.e. picking up objects). 
However, hand-crafting the gripper to hold when close to an object provides non-trivial performance on tasks, allowing MVU-R to perform selected grasping tasks. Similar findings are reported in AVDC~\citep{Ko2023LearningTA}. 
In future work, we hope to extend gripper state (possibly extracted using segmentation modules) as an additional input to our VLM and LLM modules, providing MVU-R with more holistic scene awarenss.

\end{document}